\documentclass[letterpaper]{article}

\usepackage[final]{neurips_2022}
\usepackage[utf8]{inputenc} % allow utf-8 input
\usepackage[T1]{fontenc}    % use 8-bit T1 fonts
\usepackage{hyperref}       % hyperlinks
\usepackage{url}            % simple URL typesetting
\usepackage{booktabs}       % professional-quality tables
\usepackage{amsfonts}       % blackboard math symbols
\usepackage{nicefrac}       % compact symbols for 1/2, etc.
\usepackage{microtype}      % microtypography
\usepackage{xcolor}         % colors

\usepackage{microtype}
\usepackage{graphicx}
\usepackage{subcaption}
\usepackage{color,soul} %temp for highlighting
\usepackage{algorithm,algpseudocode}

\newcommand{\errors}{\epsilon}
\newcommand{\U}{\mathcal{U}}
\newcommand{\V}{\mathcal{V}}
\newcommand{\var}{\mathbb{V}\mbox{ar}}
\newcommand{\R}{\mathbb{R}}
\newcommand{\rd}{\mathrm{d}}
\newcommand{\htild}{\tilde{h}}

\newcommand{\Lo}{{\mathcal{L}}}
\newcommand{\La}{{\mathcal{L}^*}}

\newcommand{\bp}{{\mathbf{p}}}
\newcommand{\bn}{{\mathbf{\hat{n}}}}

\usepackage{amsmath}
\usepackage{amssymb}
\usepackage{mathtools}
\usepackage{amsthm}
\usepackage{multicol}
\usepackage{dcolumn}
\usepackage{enumitem}

\usepackage{multibib}
\newcites{supp}{Supplementary References}

\usepackage[textsize=tiny]{todonotes}

\title{Adjoint-aided inference of Gaussian process driven differential equations}

\author{%
Paterne Gahungu\\ 
College of Computing\\ 
Makerere University \And Christopher W. Lanyon\\ Department of Computer Science\\ University of Sheffield
\And Mauricio A. \'Alvarez\\ Department of Computer Science\\ University of Manchester \And Engineer Bainomugisha\\College of Computing\\
 Makerere University\\ 
 \And Michael Smith\\Department of Computer Science\\ University of Sheffield\\ \And Richard D. Wilkinson\\School of Mathematical Sciences\\ 
 University of Nottingham\\}
\begin{document}

\maketitle

\begin{abstract}
  Linear systems occur throughout engineering and the sciences, most notably as differential equations. In many cases the forcing function for the system is unknown, and interest lies in using noisy observations of the system to infer the forcing, as well as other unknown parameters. In differential equations, the forcing function is an unknown function of the independent variables (typically time and space), and can be modelled as a Gaussian process (GP). In this paper we show how the adjoint of a linear system can be used to efficiently infer forcing functions modelled as GPs, using a truncated basis expansion of the GP kernel. We show how exact conjugate Bayesian inference for the truncated GP can be achieved, in many cases with substantially lower computation than would be required using MCMC methods. We demonstrate the approach on systems of both ordinary and partial differential equations, and show that the basis expansion approach approximates well the true forcing  with a modest number of basis vectors. 
Finally, we  show how to infer point estimates for the non-linear model parameters, such as the kernel length-scales, using Bayesian optimisation.
\end{abstract}

\section{Introduction}
\label{sect:intro}
Linear systems are used as models throughout the sciences and engineering, encompassing a wide range of both ordinary and partial differential equations (including the heat, wave, Schr\"{o}dinger's, Maxwell's equations etc), as well as systems of linear algebraic equations (such as eigenvalue problems).
To fix notation, let $\Lo: \U \rightarrow \V$ be a linear operator between Banach spaces $\U$ and $\V$ (i.e., complete normed spaces, \citet{kreyszig1991}) . A prototypical linear system is then of the form
\begin{equation}
\Lo u = f, \label{eqn:linsys}
\end{equation}
where $u\in\U$ is the quantity of interest being modelled, and $f\in \V$ is the {\it forcing function} of the system. Given a fully specified operator $\Lo$ and forcing function $f$ (and possibly initial and boundary conditions),  solving the system for $u$ is referred to as the {\it forward problem}. Typically, this is a computationally intensive task.
%%%Added bit
For example, consider modelling air pollution as it moves through the atmosphere. In this case, $\Lo$ may be a partial differential operator describing the advection, diffusion, and reaction of the pollution, and $f$ will be a function describing the source of the pollution at each location and time. 
The forward problem  refers to computing the concentration, $u$, given the emission sources and  will usually require the use of numerical integration methods.

%%%Removed bit
%For example, for a differential operator (such as in our PDE example in Eqs. \ref{eqn:advec} and \ref{eqn:ODE}), the solution, $u$, will be a function of one or more independent variables, and we may need to resort to numerical methods to approximate it.
%%%%%%%%%%%%

In many applications, both the linear operator $\Lo$ and the forcing $f$ may not be fully specified, and we may face the statistical task of learning $\Lo$ and $f$ from noisy observations of $u$:
\begin{equation}
z= h(u) + \errors.\label{eqn:observation} %\mbox{ where } \errors.\label{eqn:observation}
\end{equation}
Here, $z\in\R^n$ are the observations, $h$ the observation operator, $n$  the number of observations, and $\errors \in \R^n$ a zero-mean observation error. This is often referred to as the {\it inverse problem} in applied maths and statistics \citep{stuart_2010}, or sometimes as a {\it latent force model} in machine learning \citep{alvarez2009latent}. In the air pollution example this would equate to finding the distribution of pollution sources, $f$, given a set of concentration measurements $z$.

We focus on the situation where 
\begin{enumerate}[noitemsep,topsep=-4pt]
 \setlength{\itemsep}{1pt}
  \setlength{\parskip}{0pt}
  \setlength{\parsep}{0pt}
    \item $f$ is modelled as a Gaussian process (GP) when $\V$ is an infinite dimensional Banach space, or with a Gaussian distribution in the finite dimensional case. E.g., if $\Lo$ is an ordinary differential operator with independent variable $t$, then $f$ will be an unknown function of $t$.
    %\item The linear operator $\Lo$ depends (possibly non-linearly) on unknown parameters $p$. We write $\Lo_p$ to emphasise this dependence.
    \item The observation operator $h:\U \rightarrow \R^n$ is  affine. In the finite dimensional case, this implies $h(u)=Hu+c$ for some constant matrix $H$ and vector $c$, but in the infinite dimensional case (when $\U$ is a space of functions) includes pointwise evaluation of $u$, i.e. $u(x,t)$ for  some values of $x$ and $t$, and integral and derivative observations of $u$, e.g. $\int u \;{\rm d}x \rd t, \frac{\rd u}{\rd t}$ etc.
    \item The observation error $\errors$ has a Gaussian distribution. This assumption can be relaxed for maximum likelihood (ML) estimation. 
\end{enumerate}

The full specification of the statistical model is then 
\begin{equation}
\begin{aligned}
&\Lo u=f,  \qquad &z&=h(u)+\errors\label{eq:de}\\
    &f\sim GP(m(\cdot), k(\cdot, \cdot)), \qquad
& \errors&\sim \mathcal{N}_n(0, \sigma^2 I_n)
\end{aligned}
\end{equation}
% For example, $\Lo_p$ might be a linear ordinary differential operator such as Eq.~(\ref{eqn:ODE}) and $h$ might pick out the value of $u$ at time $t$.
where $m$ and $k$ are the prior mean and covariance  (kernel) functions of the GP. 
Note that the linear system in Eqs. \eqref{eq:de} may include initial and boundary conditions for differential operators. 
Our aim is to infer $f$ (and possibly $u$, $\Lo$, and $k$) given $z$ either via 

\begin{itemize}[noitemsep,topsep=-4pt]
 \setlength{\itemsep}{1pt}
  \setlength{\parskip}{0pt}
  \setlength{\parsep}{0pt}
    \item 
ML estimation, by solving the constrained optimization problem
\begin{equation}
\min_{f} \quad (z-h(u))^\top (z-h(u))+\alpha ||f||^2_\V \label{eqn:sos} \;\;\;\;\;\;\mbox{ s.t. }\; \Lo u=f\nonumber
\end{equation}
\item Bayesian inference, by computing the posterior distribution
\begin{equation}
    \pi(f, u | z) \propto \pi(z|u) \pi(u | f) \pi(f).\label{eqn:Bayes0}
\end{equation}
\end{itemize}
The prior distribution for   $f$, $\pi(f)$,  and the regularization parameter, $\alpha$,  play a similar role in the two approaches. To solve either problem numerically is likely to require many solves of the forward problem (Eq. \ref{eqn:linsys}). For example, for ML we may seek a solution using numerical optimization \citep[e.g.][]{arellano2007evaluating, troltzsch2010optimal}, whereas with Bayes, we might use an MCMC scheme \citep[e.g.][]{kopacz2009comparison,cotter2013mcmc, sengupta2016gradient,albani2021uncertainty} or a variational approach \citep{chappell2008variational}. All of these approaches require multiple solves of the forward problem. Here we develop an approach to minimize the computational cost of inference. %Our aim is to require far fewer during inference.% forward solves.  Our aim is to reduce the number of forward solves required during inference, resulting in lower computational costs.

\subsection{Contribution}
In this paper we show that implementing an adjoint of the linear system can result in much faster statistical inference. Instead of using numerical approaches to solve either the ML or Bayesian inference problem, we can do inference for $f$ at the cost of $n$ forward model solves, where $n$ is the number of data points. In many (but not all) cases, this will incur a substantially lower computational cost than competing methods, such as MCMC.  More specifically, % by developing an adjoint of the linear system, 
we show that

% show that when $f$ is modelled by a Gaussian process, th
% This paper combines \citet{rahimi2007random}'s approach for expressing Gaussian process regression as a linear regression over a set of random Fourier features and the work from \citet{hwang2019bayesian}, in which they use the adjoint to write the backward finite difference approximation as a linear model.
% By 
\begin{enumerate}
 \setlength{\itemsep}{1pt}
  \setlength{\parskip}{0pt}
  \setlength{\parsep}{0pt}
%    \item we can compute derivatives of the system with respect to $p$;
    \item if $f$ depends linearly on  parameters $q$, we can estimate $q$ or its distribution analytically, i.e. without resorting to numerical integration methods;% approximation, either via a simple equation for ML, or in a conjugate Bayesian analysis. 
    \item if we model $f$ as a Gaussian process, then by using a truncated basis expansion we can efficiently infer the posterior distribution for $f$.
\end{enumerate}

% These two benefits can radically accelerate both ML and Bayesian inference. 

The paper is structured as follows. In the next section we discuss related work before introducing adjoints in Section \ref{sect:adjoints}. We derive the main results in  Section \ref{sect:benefitsOfAdjoints}, and  in Section \ref{sect:inferenceGaussianForcing} we show how  linearizing Gaussian processes via a basis expansion reduces inference for GPs in linear systems to simple linear algebra. Finally, in Section \ref{sect:experiments} we demonstrate the approach on two linear systems: ordinary  (ODE) and partial differential equations (PDE). 

\section{Related work}

The problem defined by Eqs. \eqref{eq:de} is often referred to as a \textit{latent force model} \citep{alvarez2009latent, alvarez2013latent}. \citet{alvarez2009latent} showed how the posterior distribution (\ref{eqn:Bayes0}) can be computed by  using the integral formulation of $\Lo u = f$, i.e. $u(x)=\int G(x-v)f(v)dv$, where  $G(\cdot)$ is the Green's function associated with the differential operator $\Lo$. Due to the linearity of the integral transform, placing a GP prior over $f$ leads to a joint GP over $f$ and $u$. From this joint GP, the posterior distributions $\pi(u \mid f)$ and $\pi(u)$ can be computed in closed form\footnote{For a more detailed explanation of the Green's function method, please see the supplementary material.}.
However, in many situations, particularly for non-trivial differential equation models, the expressions for the covariances %within and between $f$ and $u$
are cumbersome and lead to the use of error functions with complex arguments or functions like the Faddeeva function that can be numerically unstable to compute. \citet{guarnizo2018latentrff} proposed representing $f$ using random Fourier features (RFFs) to reduce the number of integrations necessary to be solved analytically.  Here, rather than using Green's functions, we instead use adjoints to  write the problem as a linear model and then use a reduced-rank  Gaussian process formulation, leading to numerically stable and fast approximations to the posterior distribution.

More specifically, estimation of forcing functions in differential equation models has been extensively studied, for example, in the field of modelling atmospheric advection-diffusion \citep[e.g.][]{yee2008theory, borysiewicz2012bayesian, singh2014least, rajaona2016inference, yeo2019development} with some authors solving the inverse problem using an adjoint approach in combination  with MCMC to compute the posterior distribution \citep[e.g.][]{yee2008theory, hwang2019bayesian, luhar2020quantifying, albani2021uncertainty}. Of particular relevance is \citet{hwang2019bayesian} who use a point source model of pollution, and use the adjoint to write the backward finite difference approximation,  noting that this can be written as a linear model, where the features are conjugate fields associated with each sensor. MCMC sampling is still a limiting factor, restricting the extension of the approach to more complex situations such as time-varying pollution sources.

Other related work includes the stochastic PDE approach of \citet{lindgren2011}, in which the underlying function $u$ is modelled as a Gauss Markov random field (GMRF), which can then be formulated as a stochastic partial differential equation. This allows finite element methods to be used to efficiently compute the posterior distribution for $u$ given $z$. Similarly, \citet{hartikainen2010} exploit the link between GMRFs and dynamical systems to convert inference for $u$ to a form in which Kalman filtering methods can be used, which scale linearly with $n$. \citet{sigrist2015} focus exclusively on the advection diffusion PDE  considered in Section \ref{sect:PDE}, and use white noise for the forcing function, $f$, to create a stochastic PDE model for $u$. They use a spectral approach, solving the PDE in the Fourier domain, to develop an efficient algorithm for statistical inference. Whilst attractive, it is difficult to generalize this approach from white noise models of $f$ to correlated Gaussian process models. 
\citet{jidling2017} show how to infer statistical inference for linearly constrained systems, i.e., where $\Lo u =0$. 
These approaches all model $u$, whereas our focus is on inferring the forcing function $f$. %There is likely to be benefit in combining these approaches with ours to further accelerate inference in these models.

\section{Methods}
\label{sect:methods}
We first recap the definition of adjoints, before %and \citet{hwang2019bayesian}'s approach to convert the adjoint to a linear operation. 
 deriving our main result illustrating how they can be used to accelerate inference. We then show how using a truncated basis approximation to Gaussian processes allows us to  find the GP posterior distribution  without resorting to MCMC methods.  %provides a good introduction to adjoints. The benefit section is ,,,,, our own.. inspired by Hwang? The work on GPs follows from Rahimi and Rechts, Mercer....

\subsection{Adjoints}
\label{sect:adjoints}
Recall that  $\Lo: \U\rightarrow\V $ is a  linear mapping between Banach spaces $\U$ and $\V$. Let $\U^*$ and $\V^*$ denote the dual spaces of $\U$ and $\V$ \citep{kreyszig1991}. We can construct the adjoint to a continuous linear operator $\Lo$ as follows. Let $v^* \in \V^*$ and define $F:\U \rightarrow \R$ by
$F: u \mapsto v^*(\Lo u)$.
Then $F$ is a bounded linear functional on $\U$, i.e., $F=u^*$ for some $u^*\in\U^*$. Thus for all $v^*\in\V^*$ we've associated a unique $u^*\in\U^*$:
\begin{equation}\Lo^*: v^* \mapsto u^*=v^*\circ \Lo.\end{equation} 
We call $\La$ the {\it adjoint} of $\Lo$, and $\La$ is itself a bounded linear operator \citep{estep2004short}.  By construction, we have that for all $u\in U$ and $v^* \in \V^*$
\begin{equation}(\La v^* )(u) = v^* (\Lo u),\end{equation}
a result known as the {\it bilinear identity}.  
In the case where $\U$ and $\V$ are also real Hilbert spaces with respect to inner products $\langle \cdot, \cdot \rangle_\U$ and $\langle \cdot, \cdot \rangle_\V$, then we can identify the dual spaces with their underlying space: by the Riesz representation theorem if $v^*\in\V^*$ then there exists $v\in\V$ such that $v^*(\cdot)=\langle \cdot, v\rangle_\V$. In this case, the bilinear identity reduces to its more familiar form
\begin{align}
\langle \Lo u, v\rangle_\V &= v^*(\Lo u)=(\Lo^*v^*)(u)= \langle u, \Lo^* v\rangle_\U \label{eqn:bilinear}
\end{align}
where we now consider $\La: \V \rightarrow \U$. We only consider real vector spaces here, resulting in a symmetric inner product. 

Generally, the adjoint $\Lo^*$ will be the same type of operator as $\Lo$ (e.g. if $\Lo$ is a differential operator then $\Lo^*$ will be too), so solving an adjoint system of the form $\Lo^* v=h$ will have similar computational complexity as solving $\Lo u=f$. See \citet{estep2004short} for an introduction to adjoints, and  Section \ref{sect:experiments} for  examples of adjoint systems.

\subsection{Benefits of adjoints}
\label{sect:benefitsOfAdjoints}
How does the development of an adjoint to a linear system help us perform statistical inference for that system?  Consider  the situation where uncertainty about the unknown forcing function, $f$ in Eq.~(\ref{eqn:linsys}), can be characterized by a linear dependence upon unknown parameters $q$. That is, we can write
\begin{equation}
f(\cdot)= \sum_{m=1}^M q_m \phi_m(\cdot).\label{eqn:reducedrank}
\end{equation}
In the infinite dimensional case where $\U$ and $\V$ are  spaces of functions on some set $\mathcal{X}$, the 
$\phi_m$ will also be functions on $\mathcal{X}$. In the finite-dimensional case, the $\phi_m$ will be vectors of fixed length. 

In the situation where the observation operator (\ref{eqn:observation}) is linear, then we can write the $i^{th}$ observation as $h_i(u)=\langle h_i, u \rangle$ plus noise, for some $h_i \in \U$. Consider the $n$ different adjoint systems
$$\La v_i = h_i \mbox{ for }i=1, \ldots, n.$$
Then using the bilinear identity (\ref{eqn:bilinear}) we have that
$$h_i(u)=\langle h_i, u \rangle =\langle \La v_i, u\rangle= \langle  v_i, \Lo u\rangle= \langle  v_i, f\rangle,
$$
i.e., the $i^{th}$ observation is the inner product between the unknown forcing function $f$ and the solution of the $i^{th}$ adjoint system. At first, the introduction of the adjoint doesn't appear to have helped. To evaluate the likelihood (or sum of squares) we have gone from needing a single solve of the forward problem, to requiring the solution to $n$ adjoint systems. %: an $n$-fold increase in computational cost. 
The benefit arises if we now use the assumption of a linear dependence upon the parameters, and linearity of inner products:
$$h_i(u) = \langle  v_i, \sum_{m=1}^M q_m \phi_m \rangle=\sum_{m=1}^M q_m \langle  v_i,  \phi_m\rangle.$$
The complete observation vector $z$ can then be written as
\begin{equation}
z = \begin{pmatrix} \langle v_1, \phi_1 \rangle & \ldots & \langle v_1, \phi_M \rangle\\
\vdots&&\vdots\\
\langle v_n, \phi_1 \rangle & \ldots & \langle v_n, \phi_M \rangle
\end{pmatrix}
\begin{pmatrix} q_1\\
\\
q_M\end{pmatrix}+\errors
=\Phi q+\errors%\nonumber
\label{eqn:Phi}
\end{equation}
where $q\in\mathbb{R}^M$ is the parameter vector, and 
$\Phi\in \mathbb{R}^{n\times M}$ is the matrix of inner products between the $n$ adjoint solutions and $M$ basis vectors. 

This can be recognized as a  linear model. Thus, standard results can be used to compute the least squares/ML and Bayesian estimators. For ML, the minimum of $S(q) = (z-h(u))^\top (z-h(u))$ subject to $\Lo u=f$, 
is obtained at
$\hat{q} = (\Phi^\top \Phi)^{-1} \Phi^\top z$ with
$\var(\hat{q}) = \sigma^2(\Phi^\top \Phi)^{-1}$ in the case where the observation errors $\errors_i$ are uncorrelated and homoscedastic with variance $\sigma^2$. Standard results can be used from regularized least squares if we need to regularize $q$. 

In a Bayesian setting, if we assume {\it a priori} that  $q\sim \mathcal{N}_M(\mu_0, \Sigma_0)$, then the posterior for $q$ given $z$ (and other parameters) is 
\begin{equation}
q \mid z \sim \mathcal{N}_M(\mu_n, \Sigma_n)\label{eqn:Bayes1}
\end{equation}
where
\begin{equation}
\mu_n = \Sigma_n \left(\frac{1}{\sigma^2}\Phi^\top z +\Sigma_0^{-1} \mu_0\right), \;\; \Sigma_n=\left(\frac{1}{\sigma^2}\Phi^\top \Phi + \Sigma_0^{-1}\right)^{-1}.\label{eqn:Bayes2}
\end{equation}
See, e.g., \citet{o2004kendall}. Note that the solutions of the adjoint equation, $v_i$, are present in the posterior distribution of $q$ in the matrix $\Phi$.

%When will using an adjoint reduce computational expense?

\subsection{Inference of Gaussian forcing functions}
\label{sect:inferenceGaussianForcing}

We now consider the case where the forcing function is given a Gaussian process prior distribution:
\begin{eqnarray}
f(\cdot)\sim GP(m(\cdot),k(\cdot,\cdot)),
\end{eqnarray}
where $m(\cdot)$ and $k(\cdot,\cdot)$ are the prior mean and covariance functions respectively \citep{rasmussen2003gaussian}. Our approach is to use a reduced-rank representation of the GP, as in Eq.~(\ref{eqn:reducedrank}), derived by truncating an expansion for $k$ of the form
\begin{eqnarray}
	k(x,x')=\sum_{m=1}^{\infty}\phi_m(x)\phi_m(x').
	\label{alternativek}
\end{eqnarray}
There are many possible choices for the basis vectors $\phi_i(\cdot)$, including the Karhunen-Lo\`eve (KL) \citep{deheuvels2008karhunen}, Laplacian \citep{solin2020hilbert,coveney2020gaussian,borovitskiy2020mat}, and  random Fourier feature (RFF) \citep{rahimi2007random} expansions.
The Karhunen Lo\`eve basis is formed by finding the spectral expansion of the integral operator defined in Mercer's theorem: 
$$T_k f(x) = \int_\mathcal{X} k(x,x') f(x') \rd x'.$$
In the case where $\dim(x)$ is small (such as the ODE example below), the eigenfunctions of $T_k$ are easy to compute either analytically \citep[Section 4.3.2, p116]{rasmussen2003gaussian} or numerically \citep{greengard2021efficient}. Using the eigenfunctions of $T_k$ gives the $L^2$-optimal approximation \citep{kosambi2016statistics}, but the eigenfunctions can be difficult to compute even in three dimensions. A simpler approach that extends easily to higher dimensional problems, is to use RFFs  \citep{rahimi2007random}. This relies upon Bochner's theorem, which can be used to express stationary kernels  $k(x,x'):=k(x-x')$ as the Fourier transform of a positive measure $p$
\begin{eqnarray*}
		k(x-x')=\int \exp(-iw(x-x'))\rd p(w).
\end{eqnarray*}
% If we use an isotropic exponentiated quadratic (EQ) kernel, 
% \begin{equation}k(x-x')= \tau^2 \exp(-\frac{1}{2\lambda^2}(x-x')^\top(x-x'))\label{eqn:EQ}
% \end{equation}
% then the measure corresponds to a multivariate Gaussian. % $p(w)\sim \mathcal{N}_d(0, I)$.
% Using results from \citet{rahimi2007random}, we can approximate the Gaussian process with Fourier feature vectors:
% \begin{equation}
% \phi_i(x) = \sqrt{\frac{2\tau^2}{M}} \cos( \frac{1}{\lambda}w_i^\top x+b_i)\label{eqn:rff}
% \end{equation}
% where
% $w_i\sim N(0,1)$ and 
% $b_i \sim U(0, 2\pi)$. The extension to anisotropic kernels is straight-forward.
If we use an isotropic exponentiated quadratic (EQ) kernel $k$, then the measure $p$ corresponds to a multivariate Gaussian, and % $p(w)\sim \mathcal{N}_d(0, I)$. 
\citet{rahimi2007random} give expressions for the Fourier feature basis vectors that can be used to approximate the Gaussian process:
\begin{equation}
k(x-x')= \tau^2 \exp(-\frac{1}{2\lambda^2}(x-x')^\top(x-x')) \quad \Rightarrow \quad
\phi_m(x) = \sqrt{\frac{2\tau^2}{M}} \cos( \frac{1}{\lambda}w_m^\top x+b_m)\label{eqn:rff}
\end{equation}
where
$w_m\sim N(0,I)$ and 
$b_m \sim U(0, 2\pi)$. Substituting this $\phi_m$ into equation \ref{eqn:reducedrank} allows us to write the forcing function $f$ as a truncated Gaussian process with an EQ kernel (as such the true Gaussian process is never calculated). The extension to anisotropic kernels is straight-forward.

Although the RFF expansion will require more basis vectors (a larger $M$) than  the KL expansion to achieve the same accuracy, the computational complexity of our adjoint approach is dominated by the number of adjoint solves, not the number of features (which only affects the cost of computing the relatively low-cost Eq.~\ref{alternativek}), and so including more terms has a minor effect on overall computational cost.  

\subsection{Time complexity of Algorithm 1} 
Let $G$ be the number of grid elements, $n$ the number of observations, and $M$ the number of features. There are five stages in our algorithm:
\begin{enumerate} 
\item Solving the $n$ adjoint systems, which requires $\mathcal{O}(Gn)$ operations. 
\item Computing each basis vector, $\phi$, over the grid for each feature  requires $\mathcal{O}(GM)$ operations. 
\item Computing the  matrix $\Phi$, requires $\mathcal{O}(GMn)$ operations. 
\item Finding $\Phi^\top \Phi$ requires $\mathcal{O}(nM^2)$ operations. 
\item Finally solving the matrix inverse will require $\mathcal{O}(M^3)$ operations.
\end{enumerate}
This results in an algorithm that scales linearly in the number of data points, $n$. Empirically, for the problems we've experimented with, we found computing $\phi$ over the large grid was the most time consuming step. The overall time complexity is $\mathcal{O}(GMn) + \mathcal{O}(nM^2) + \mathcal{O}(M^3)$, but note that the constants associated with each term are important in most problems.

\section{Experiments}
\label{sect:experiments}
% \subsection{Linear algebraic system}
% Possibly...

\subsection{Ordinary differential equation (ODE)}
\label{sect:ODE}
Consider the non-homogeneous linear ODE:
\begin{eqnarray}
	p_2\frac{\rd^2u}{\rd t^2}+p_1\frac{\rd u}{\rd t}+p_0 u=f(t) \label{eqn:ODE}
\end{eqnarray}
on the domain $[0, T]$ with initial conditions $u(0)=u'(0)=0$. The right hand side, $f(t)$, is the  unknown forcing function that we wish to estimate, and $p_0, p_1, p_2$ are parameters in the linear operator. We model $f$ as a zero-mean GP with an EQ kernel (Eq.~\ref{eqn:rff}).%\quad \mbox{with} \quad k(t,t')=\tau \exp((t-t')^2/\lambda).$
We assume observations are obtained as noisy averages over short time windows:
\begin{equation}
h_i(u) = \int_{t_i}^{t_i+\Delta t} \frac{1}{\Delta t} u(t) \rd t=\langle u, \htild_i\rangle \label{eqn:odenoise}
\end{equation}
where $\htild_i$ is the indicator function $\htild_i(t) = \mathbb{I}_{[t_i, t_i+\Delta t]}(t)$. This includes direct measurements of $u(t_i)$ if we  set $\htild_i(t) = \delta(t-t_i)$, the Dirac delta function. To generate synthetic data, we  simulate a realization $f$ from the GP model, solve Eq.~(\ref{eqn:ODE}) for $u(t)$, and then simulate $n$  observations from Eq.~(\ref{eqn:odenoise}) with $T=1$, $p_2=0.5, p_1=1, p_0=5$, $\Delta t = \frac{T}{n}$, $t_i = \frac{iT}{n}$, adding zero-mean Gaussian noise with standard deviation $\sigma=0.1$. For simplicity, we solve the ODE with a simple forward Euler approximation, but higher order schemes  can and should be used in real applications. We approximate the GP using Eq.~(\ref{eqn:reducedrank}), using $200$ RFFs generated using  Eq.~(\ref{eqn:rff}) with $\lambda=\sqrt{0.6}$ and $ \tau^2=4$. The linear operator in this case is 
$$\Lo u =  \left(p_2 \frac{\rd^2}{\rd t^2} + p_1 \frac{\rd}{\rd t}  + p_0\right)u.$$
 To derive the adjoint operator we use the bilinear identity (Eq.~\ref{eqn:bilinear}), and integrate by parts twice:
%  \begin{align*}
% \langle \Lo u, v \rangle &= \int_0^T (\Lo u(t)) v(t) \rd t= \int_0^T (p_2 \ddot u + p_1 \dot u  + p_0 u)v \rd t\\
% %&=\left[p_2\dot u v+p_1 u v\right]_0^T+\int_0^T -p_2\dot u\dot v  - p_1 u\dot v + p_0 u v \rd t\\
% %&= \left[p_2(\dot u v+u\dot v)+p_1 u v\right]_0^T
% %+\int_0^T p_2 u\ddot v    - p_1 u\dot v + uv \rd t\\
% &= \int_0^T (p_2\ddot v - p_1\dot v+v)u \rd t   =\langle u, \La v \rangle 
% \end{align*}
\begin{align*}
\langle \Lo u, v \rangle = \int_0^T (\Lo u) v \rd t= \int_0^T (p_2 \ddot u + p_1 \dot u  + p_0 u)v \rd t = \int_0^T (p_2\ddot v - p_1\dot v+v)u \rd t   =\langle u, \La v \rangle 
\end{align*}
when $v(T)=\dot v(T)=0$. So the adjoint of $\Lo$ is
$$\La v = \left(p_2 \frac{\rd^2}{\rd t^2} - p_1 \frac{\rd}{\rd t}  + p_0\right)v.$$
Note that rather than an initial condition, the adjoint  has a final condition: to solve the system we have to integrate backwards in time from $t=T$ to $t=0$.
See Algorithm \ref{alg:inference}.
\begin{algorithm}
\caption{Computing the posterior distribution of $q$}\label{alg:inference}
\begin{algorithmic}
\For{$i=1\ldots n$}
\State Solve adjoint system $\La v_i = \htild_i$ with appropriate final and boundary conditions.%final condition $v_i(T)=\dot v_i(0)=0$.
\EndFor
\For {$m=1\ldots M$}
\State Sample an RFF basis vector $\phi_m$ using Eq.~(\ref{eqn:rff}). 
    \For {$i=1 \ldots n$}
        \State Compute $[\Phi]_{im} = \langle v_i, \phi_m\rangle$.
    \EndFor
\EndFor
\State Compute the posterior distribution for $q$ using Eqs. (\ref{eqn:Bayes1}) and  (\ref{eqn:Bayes2}).
\end{algorithmic}
\end{algorithm}

\begin{figure}[ht!]
\begin{centering}
\begin{subfigure}{.33\linewidth}
    \includegraphics[width=\textwidth]{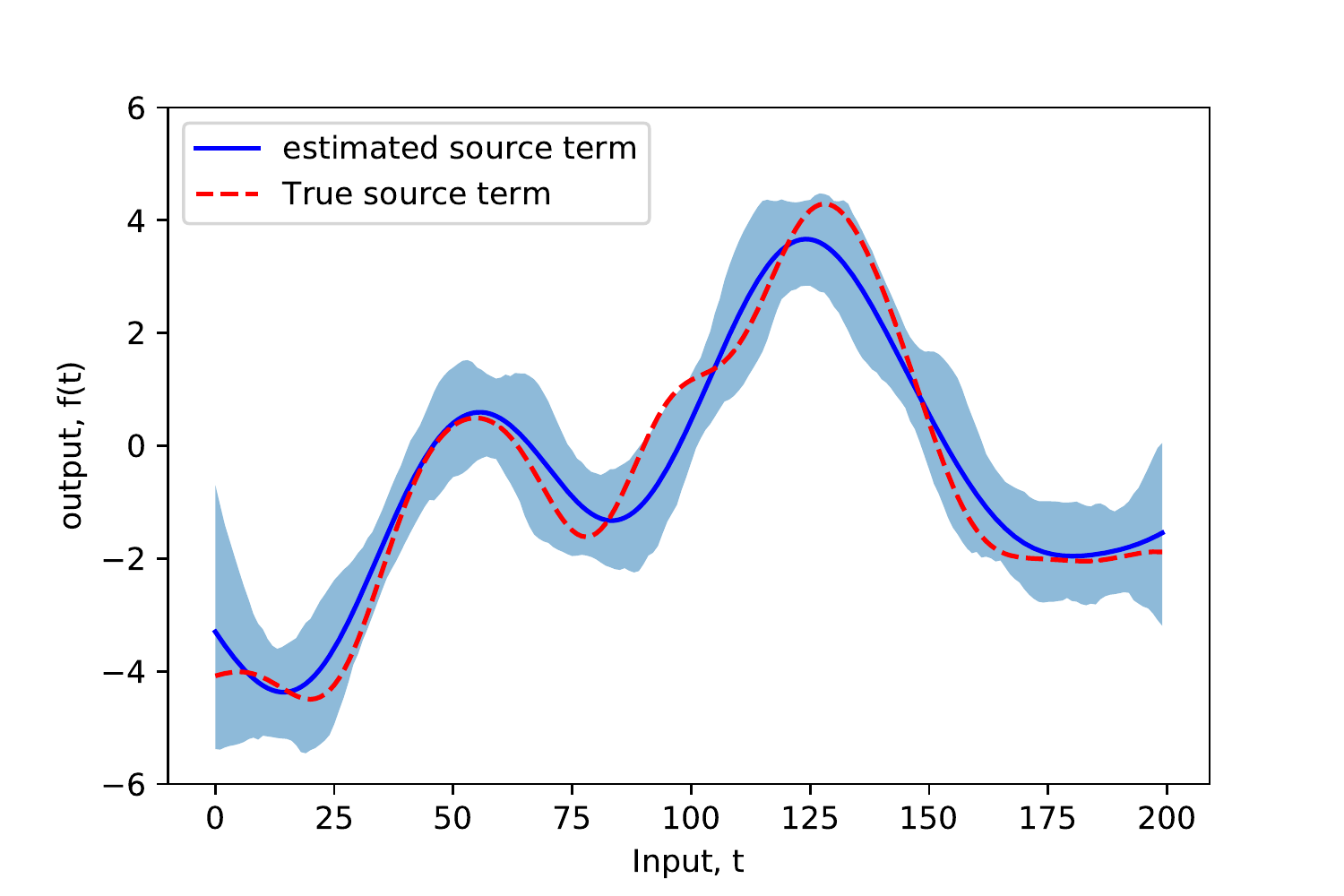}
    \end{subfigure}%
\begin{subfigure}{.33\linewidth}
    \includegraphics[width=\textwidth]{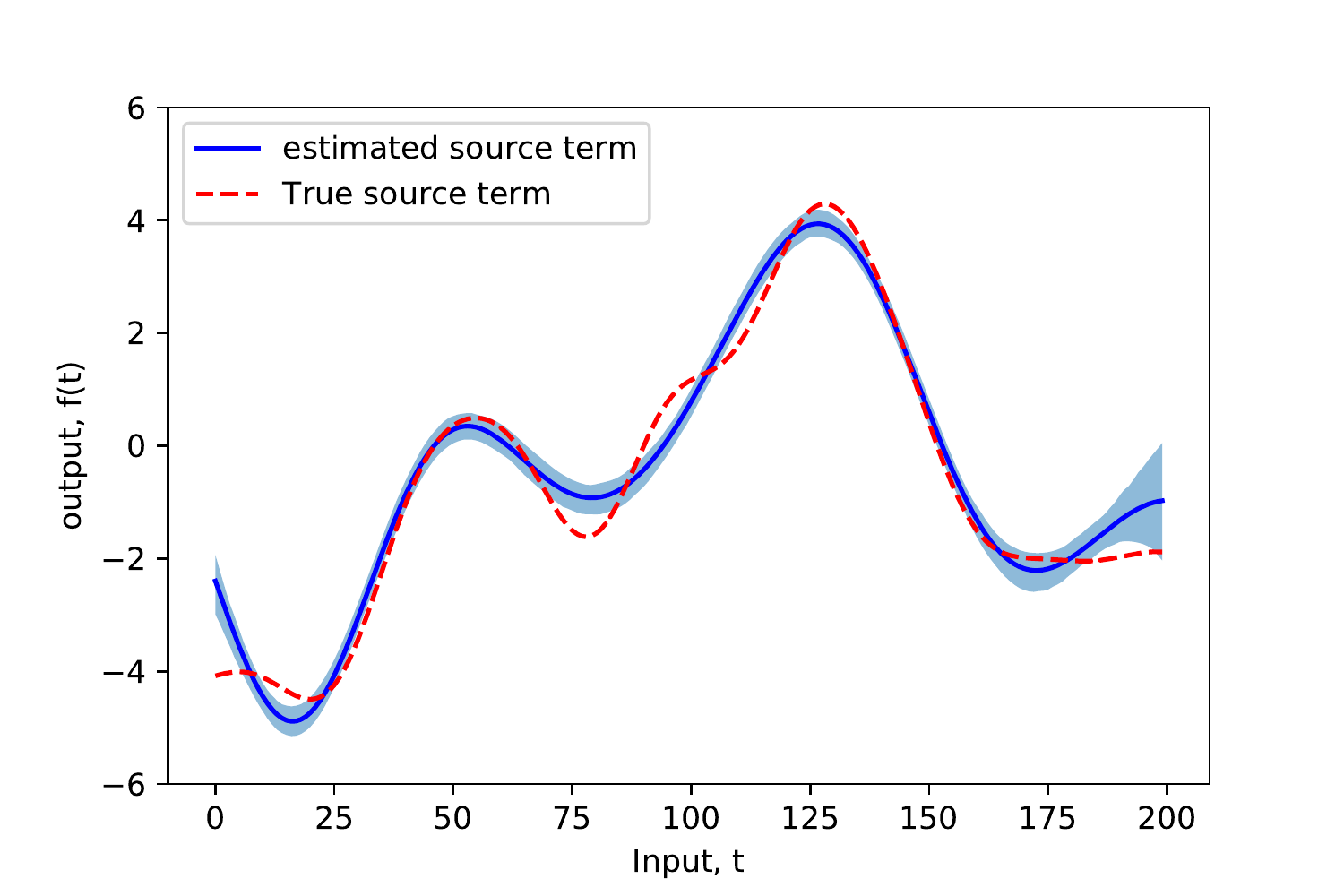}
    \end{subfigure}%
\begin{subfigure}{.33\linewidth}
    \includegraphics[width=\textwidth]{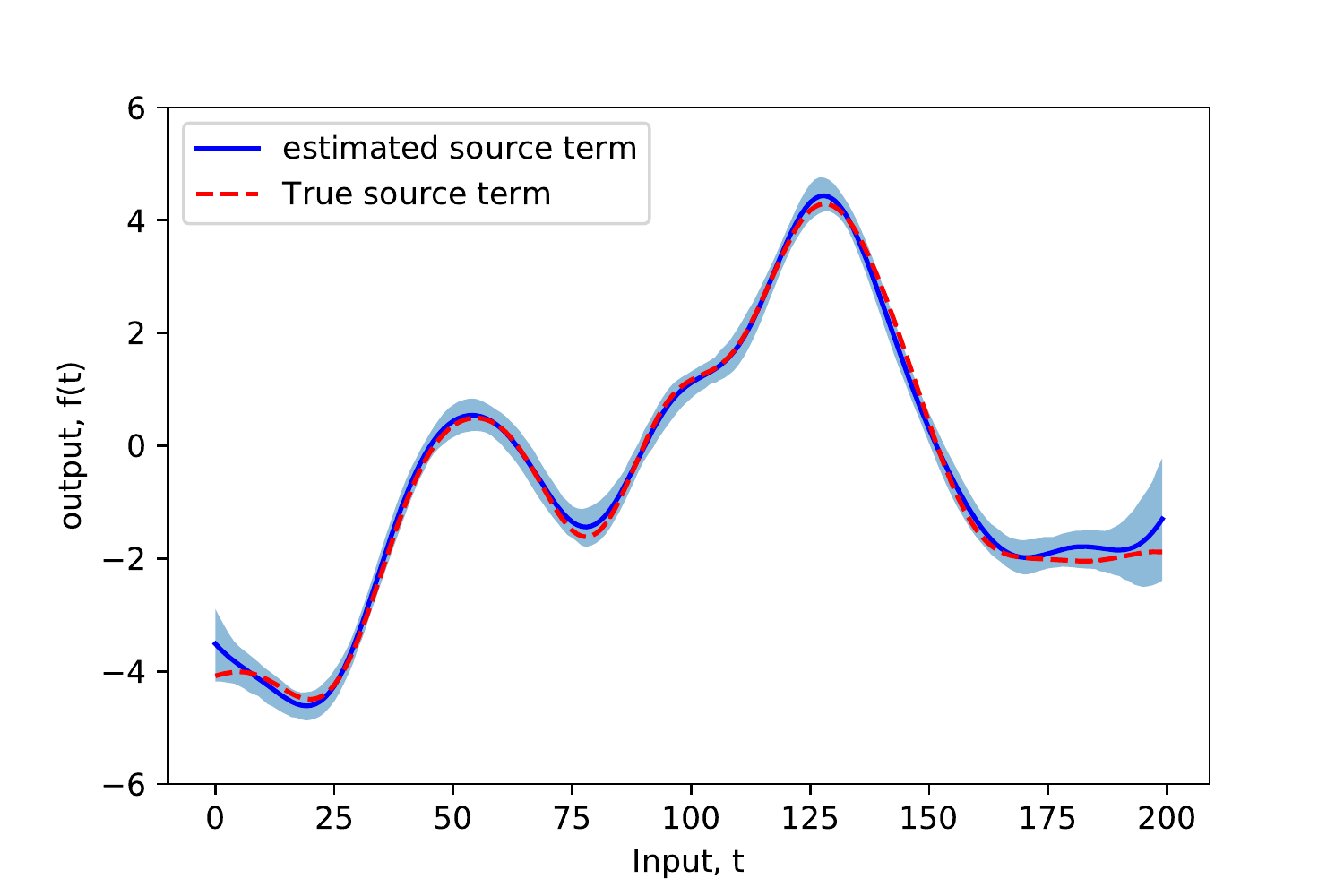}\par
    \end{subfigure}%
\end{centering}
\caption{Posterior distribution for the unknown forcing function (with 95\% credibility interval). True forcing in red. The number of training points and features are, (left) $n=10$ and $M=100$; (middle) $n=100$ and $M=10$; (right) $n=100$ and $M=100$. The overconfident and wrong posterior when $M=10$ is a consequence of the model being heavily misspecified.}\label{fig2}
\end{figure}

Fig.~\ref{fig2} shows the effects of the number of training points $n$ and the number of features $M$ on the posterior distribution of the forcing function. As expected, more data results in a more confident posterior. Note though the danger of using too few features: with $M=10$ the approximation of $f$ has limited expressive power and cannot capture the true form of $f$, i.e., the model is heavily misspecified. This can result in the uncertainty collapsing upon the most likely, but wrong, value. This can be difficult to spot, so users should check the sensitivity of the posterior with respect to $M$ (which can be done with no additional forward solves).

In the supplementary material we provide a comparison between the adjoint method, the Green's function methods as in \citet{alvarez2009latent} and \citet{guarnizo2018latentrff}, and a vanilla Gaussian process when applied to this ODE problem. 

% FIG 2 WAS HERE RDW

% \begin{figure}[ht]
% \begin{centering}
% \begin{multicols}{2}
% \begin{subfigure}{.45\textwidth}
%     \includegraphics[width=\linewidth]{Figures/NumberFeature10.pdf}
%     \caption{} https://www.overleaf.com/project/61b718506aca585ff8a1408d
% \label{fig3:a}
%     \end{subfigure}%
%     \end{multicols}
% \begin{multicols}{2}
% \begin{subfigure}{.45\textwidth}
%     \includegraphics[width=\linewidth]{Figures/NumberFeature150.pdf}
%     \caption{} 
% \label{fig3:b}
%     \end{subfigure}%
% \end{multicols}
% \end{centering}
% \caption{True and inferred forcing functions using $M=10$ (a), and $M=150$ (b) features in the GP expansion (Eq.~\ref{eqn:reducedrank}). The over confident and wrong posterior in (a) is a consequence of the model being heavily misspecified when using only $10$ features.}
% \end{figure}

\subsection{Partial differential equation (PDE)}\label{sect:PDE}

We now demonstrate the approach on a PDE in which there are two spatial variables, $x\in \mathcal{X}\subset \mathbb{R}^2$, and time, $t\in [0,T]\subset \mathbb{R}$, so that the solution $u\equiv u(x,t)$ is a function of three independent variables. We consider the advection-diffusion equation
\begin{equation}
\frac{\partial u}{\partial t} + p_1\cdot\nabla u -\nabla \cdot (p_2 \nabla u)=f \mbox{ in } \mathcal{X}\times [0,T]
\label{eqn:advec}
\end{equation}
with initial and boundary conditions
$u(x,0)=0$ for  $x\in \mathcal{X}$ and $\nabla_{n}u=0$ for $x\in\partial \mathcal{X}$. 
% with the following initial and boundary conditions:
% \begin{equation}
% u(x,0)&=0 \mbox{ for } x\in \mathcal{X}, \qquad
% \nabla_{n}u&=0    \mbox{  for }x\in\partial \mathcal{X}.
% \label{eqn:PDEIC}
% %\label{eqn:PDEBC}
% \end{equation}
Here, the unknown forcing function $f\equiv f(x,t)$ is a function of space and time, and we model it as a zero-mean Gaussian process, $f(x,t) \sim GP(0,k((x,t),(x',t'))$ with EQ kernel $k$ (Eq.~\ref{eqn:rff}). We use an RFF approximation to $k$ with random weights  $w_m\sim \mathcal{N}_3(0, I)$ and $b_m \sim U(0,2\pi)$.

Observations are assumed to arise from {\it sensors} which take an average of $u(x,t)$  over a small spatial and temporal window $\mathcal{R}_i \times \mathcal{T}_i\subset \mathcal{X}\times [0,T]$:
\begin{equation*}
    z_i = \langle u, \htild_i\rangle + e_i \quad
    \mbox{with}\quad \htild_i = \begin{cases}
\frac{1}{|\mathcal{R}_i|.|\mathcal{T}_i|} & \mbox{ if } x\in \mathcal{R}_i \mbox{ and } t\in \mathcal{T}_i \\
0 &\mbox{ otherwise}.
    \end{cases}
\end{equation*}
% \begin{align}
%     z_i &= \langle u, \htild_i\rangle + e_i\\
%     \mbox{with}\quad \htild_i &= \begin{cases}
% \frac{1}{|\mathcal{R}_i|.|\tau_i|} & \mbox{ if } x\in \mathcal{R}_i \mbox{ and } t\in \mathcal{T}_i \\
% 0 &\mbox{ otherwise}.
%     \end{cases}
% \end{align}
The adjoint of the linear operator $\Lo u = \frac{\partial u}{\partial t} + p_1.\nabla u -\nabla \cdot (p_2 \nabla u)$ 
is
\begin{equation*}
\La v=-\frac{\partial v}{\partial t}-p_1.\nabla v-\nabla \cdot(p_2\nabla v).
\end{equation*}
%This  can be derived by integrating by parts as for the ODE, but for partial differential operators, we rely  upon Green's theorem to do so; 
Our adjoint-aided approach then requires the solution of
\begin{equation*}
\La v_i = \htild_i \mbox{ in } \mathcal{X} \times [0,T] \;\; \text{for $i = 1, \ldots, n$.}
\end{equation*}
%where the initial and boundary conditions on the original system become \emph{final} and boundary conditions on the adjoint system:
%where the initial conditions on the original system become \emph{final} conditions on the adjoint system,
The final and boundary conditions on the adjoint system are
\begin{equation*}
v_i(x,T)=0  \mbox{ for }x\in \mathcal{X},\quad \mbox{ and }\quad 
p_1v_i(x,t)+p_2\nabla v(x,t)=0 \mbox{ for } x\in \partial \Omega \mbox{ and } t\in [0,T].
\end{equation*}
For details of the adjoint derivation see the supplementary material and \citet{estep2004short}. 

Inference then proceeds as before. After solving the $n$ adjoint systems, we compute the inner product of these solutions with the RFF basis vectors to form the matrix $\Phi$ (Eq.~\ref{eqn:Phi}).  We can then use Eq.~(\ref{eqn:Bayes2}) to compute the posterior with minimal additional computational cost.

Data was  simulated on the spatial domain $\mathcal{X}=[0,10]^2$ for $t\in [0,10]$ by first randomly generating a forcing function $f(x,t)$ (generated from a GP using an EQ kernel with $\lambda=2$, $\tau^2=2$), and then solving the forward problem (Eq. \ref{eqn:advec}) to find $u(x,t)$ using PDE parameters $p_1=(0.4,0.4)$ and $p_2=0.01$. We generate $n$ observations using sensors that record averages over short time windows equally spaced across the domain $[0,10]$ at the locations  shown in Fig.~\ref{fig:varT5}. 
Zero-mean Gaussian distributed noise is added to the true sensor readings  with standard deviation $\sigma=0.05$ (note that this is relatively small compared to the signal, which can often create problems for sampling methods). 
We then use Algorithm \ref{alg:inference} to calculate the posterior distribution for $q$, hence giving the posterior for $f$. By sampling forcing functions from this posterior and simulating forward, we can evaluate the posterior predictive accuracy of the model.

To validate the posterior estimates from the adjoint method, we also used MCMC to  compute the posterior distribution for the PDE model using just $M=10$ RFFs. Table \ref{tab:MHvAdj} shows the posterior mean and variance of the first 5 $q$ parameters determined using both methods, which can be seen to be in close agreement. Fig.~\ref{fig:MCMC} in the supplementary material shows the trace plots. We used a batch-update random-walk Metropolis-Hastings (MH) sampler, which failed to converge (after an hour of computation) when using a larger number of features. Even with $M=10$, we still required 10,000s of forward model evaluations to reach convergence (whereas in comparison, the adjoint method required just 75 forward model solves  in this case).

%the problem, $u(x,t)$, was first calculated using a randomly generated source term. Each sensor in the array then takes a reading at times $t_1,t_2,t_3,...,t_{end}$, ($t_i \in [0,T] \forall i $). The sensor locations and reading times dictate the upper and lower bounds of the filter function, $h$. 

%section{PDE model results}

% As a first example, an artificial source was generated with 1000 RFFs to approximate a GP with an exponentiated quadratic kernel ($\lambda=2$, $\tau^2=2$), this source function is shown in \ref{fig:SourceInferEx}. The source was then used to simulate the PDE problem and observations were generated using 4 and 16 sensor arrays (the source locations can be seen in Figure \ref{fig:sensorVar}). The adjoint method was then used with 200 RFFs to infer the original source, $q$. The inferred mean values of $q$ are shown in Figures \ref{fig:SourceInferEx4} and \ref{fig:SourceInferEx16}.  It can be seen that the training data generated with a higher sensor density results in a better estimation of key features of the original source. 

\begin{figure}[]
\begin{center}
\includegraphics[height=3.5cm]{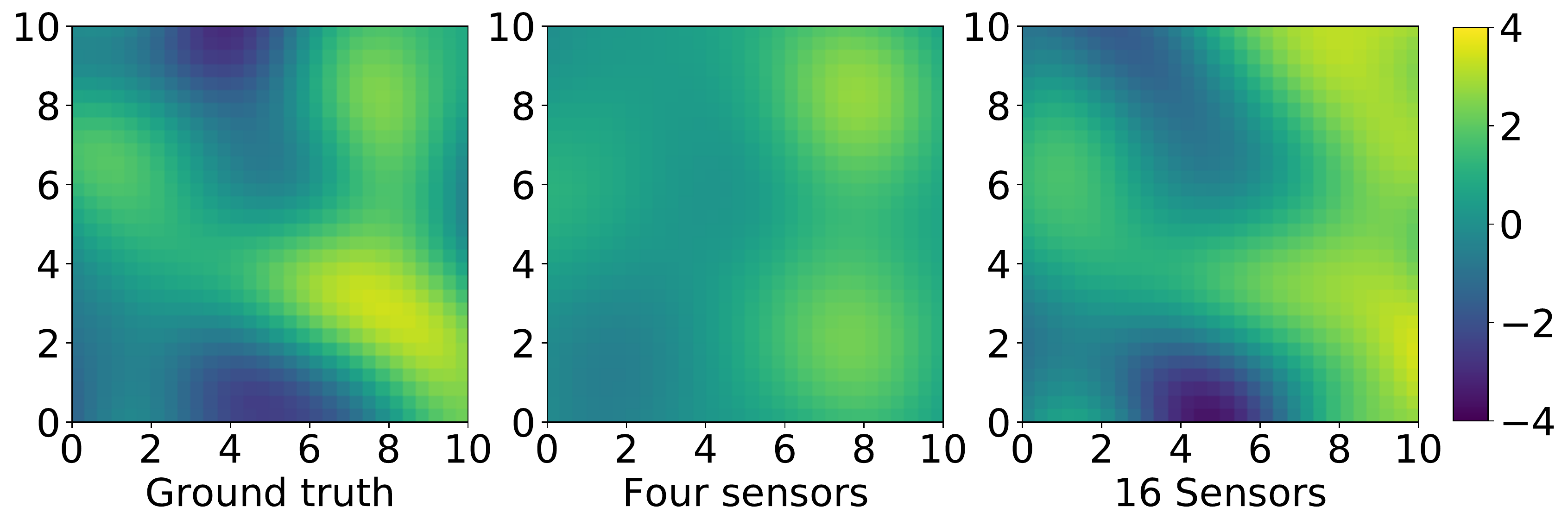}
\end{center}
\caption{Spatial maps of the forcing function at a single slice: $f(x,5)$. Left shows ground truth, middle shows the posterior mean with 4 sensors, right shows the posterior mean using 16 sensors.}
\label{fig:SourceInferT5}
\end{figure}

\begin{figure}[]
\begin{center}
\includegraphics[height=3.5cm]{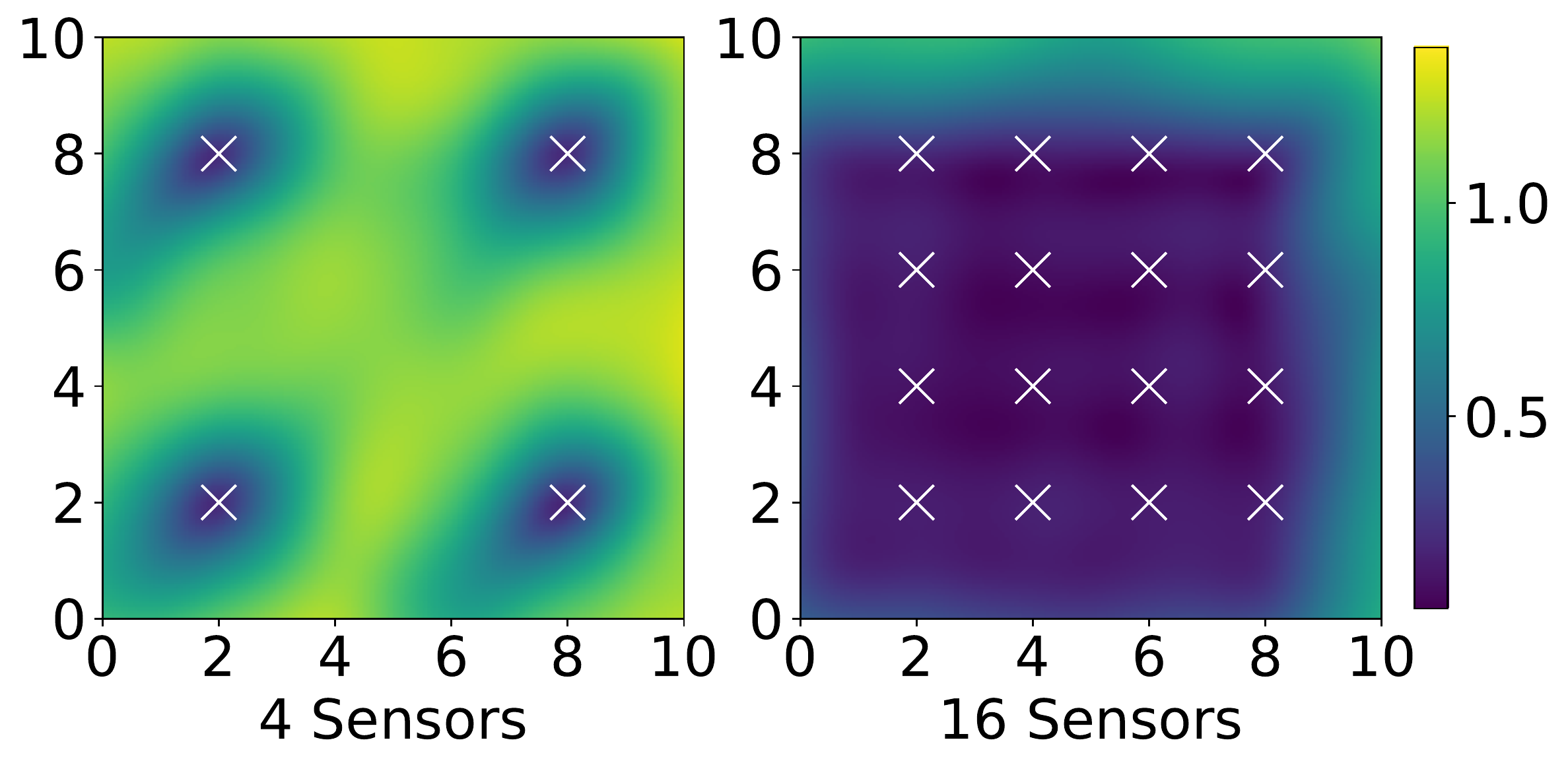}
\end{center}
\caption{Posterior standard deviation for the examples shown in Fig.~\ref{fig:SourceInferT5}. The left image is for the case where 4 sensors are used, and the right for 16 sensors. Sensor locations are shown as white crosses.}
\label{fig:varT5}
\end{figure}

To illustrate the effects of sensor density, observations were generated for five time windows using arrays of 4 and 16 sensors at the locations shown in Fig.~\ref{fig:varT5}, i.e., a total of either $n=20$ or $n=80$ observations. The ground truth forcing and inference results (with $M=200$) are shown in Fig.~\ref{fig:SourceInferT5}. As expected, more sensors results in improved estimates. 
Fig. \ref{fig:varT5} shows the posterior standard deviation for $f$ in the 4 and 16 sensor cases. Here  advection occurs  parallel to the $y=x$ line as $p_1=(0.4, 0.4)$ (i.e., as if there is wind blowing from the  south west). We can see that standard deviation is smallest {\it upwind} of the sensors, with more uncertainty {\it downwind}, as expected.

 \begin{table}
 \centering
  \caption{Posterior mean and standard deviation of the $q$ parameters estimated by MCMC and the adjoint method.}
 \begin{tabular}{lccccc}
 & $q_1$ & $q_2$ & $q_3$ & $q_4$ & $q_5$ \\
 \hline
MCMC & -3.62 (0.02) & -0.64 (0.01) & 1.68 (0.01) & -0.09 (0.01) & 4.31 (0.02) \\
Adjoint & -3.62 (0.02) & -0.64 (0.01) & 1.68 (0.01) & -0.09 (0.01) & 4.31 (0.02) \\
 \end{tabular}

\label{tab:MHvAdj}
\end{table}

To investigate the effects of varying feature and sensor numbers we performed a posterior predictive check using held-out data and used Monte Carlo estimation to calculate the posterior predictive mean squared error (MSE). Table \ref{tab:FeatSensDep3} gives the MSE as the number of sensors and RFFs vary. As both increase, so does the accuracy of our estimates. In general, the accuracy depends upon a variety of factors, including the PDE parameters (ratio of diffusion to advection), kernel parameters (decreasing lengthscale makes the problem more challenging), and sensor locations. The speed and efficiency of the proposed adjoint-aided approach allows us to investigate these effects in a way that would not be possible if we were using  MCMC (as each estimate of the posterior requires tens of thousands of simulator evaluations, rather than just the $n$ evaluations required for the adjoint-aided approach).

%It should be noted that the accuracy of the model is dependent on a variety of factors including the kernel lengthscale, the PDE parameters and the locations of the sensors. Therefore no general conclusions about required numbers of features or sensors should be drawn from this analysis.

\begin{table*}
\scriptsize
\centering
\caption{%The posterior predictive MSE as a function of the number of sensors and RFFs. The ground truth was generated from a GP with $\lambda=2$ and $\tau^2=2$, and PDE parameters $p_1=(0.01,0.01)$ and $p_2=0.01$. Here the median MSE from 10 samples is reported alongside the 95\% confidence interval. We also report the MSE for a Metropolis Hastings (MH) algorithm with 25 sensors (5 samples). Unfortunately for 50 or more features, the MH algorithm does not converge after 20000 iterations.
The median MSE as a function of number of sensors and RFFs. The ground truth was generated using a GP with $\lambda=2$, $\tau^2=2$ and PDE parameters $p_1=(0.01,0.01)$, $p_2=0.01$. The MH algorithm did not converge after 20,000 iterations for 50 or more RFFs. The numbers  brackets are the 95\% confidence intervals computed from 10 repeated experiments (for the adjoint approach) and 5 (for MH).} 
\begin{tabular}{lllllll}
\multicolumn{1}{l}{Sensors} & \multicolumn{6}{c}{Features}                                                                                        \\
\multicolumn{1}{l}{}        & 10                 & 50                & 100              & 200              & 300              & 500               \\ 
\hline
1                           & 3.42 (2.82,4.39)   & 3.27 (3.13,3.38)  & 3.24 (3.10,3.37) & 3.27 (3.17,3.44) & 3.24 (3.12,3.36) & 3.27 (3.17,3.35)  \\
4                           & 7.12 (1.57,28.81) & 2.39 (2.06,2.62)  & 2.41 (2.13,2.60) & 2.45 (2.32,2.57) & 2.50 (2.41,2.69) & 2.53 (2.32,2.60)  \\
9                           & 2.38 (1.41,4.40)  & 2.12 (1.48,3.98)  & 1.70 (1.49,2.07) & 1.48 (1.40,1.72) & 1.47 (1.32,1.61) & 1.45 (1.40,1.50)  \\
16                          & 1.73 (1.23,3.28)   & 3.99 (2.32,10.90) & 2.18 (1.72,3.54) & 1.3 (1.02,1.68) & 1.12 (0.98,1.37) & 1.12 (1.02,1.21)  \\
25                          & 1.35 (1.19,3.09)   & 8.93 (4.92,39.86) & 4.36 (2.53,8.20) & 1.86 (1.43,2.75) & 1.35 (1.07,1.81) & 1.05 (0.89,1.45)  \\
25 (MH)                   &   3.27 (1.73,6.12)                 &        -           &        -          &  -               &        -          &       -        
\end{tabular}
\label{tab:FeatSensDep3}
\end{table*}

\iffalse
\begin{figure}[ht]
\includegraphics[width=\columnwidth]{Figures/MSE.pdf}
\caption{The  posterior predictive mean squared error (MSE) as a function of the  
number of sensors and RFFs. The ground truth was generated from a GP with $\lambda=2$ and $\tau^2=2$, and using advection-diffusion parameters $p_1=(0.04,0.04)$ and $p_2=0.01$. }
\label{fig:FeatSensDep3}
\end{figure}
\fi

%%%%%%%%% REWRITE BELOW WITHOUT IMAGE

Although not our focus here, we note that we can infer the remaining parameters $p$ (PDE) and $(\tau^2, \lambda)$ (GP hyperparameters) in a variety of ways. For example, using Bayesian optimization  we can find the maximum likelihood estimates of these parameters with relatively minimal computational cost. See the supplementary material for details and examples. 

In the supplementary material we include some examples of the effect of varying the Gaussian process lengthscale on inference quality, an example in which we apply our method to an advection diffusion problem using the Roundhill II dataset \citep{cramer1957field} and apply our method to shift operators (a non-differential linear operator).   

\subsection{Real time cost analysis}
\label{sect:realtime}
We compared the run time of the adjoint method to a basic MH MCMC algorithm (recorded on a laptop with 16GB RAM and an Intel i7-1065G7 CPU @ 1.50 GHz). Exact inference of the posterior of $f$ in the ODE model using the adjoint method on a $100$ element temporal grid with 100 observations using 100 RFFs took 569 $\pm$ 72.5 ms. This is approximately the time it took to perform a single iteration of the MH algorithm.
Inference of the posterior of $f$ in the PDE model using the adjoint method on a $50\times 30 \times 30$ spatiotemporal grid with 100 observations using 100 RFFs took 3.32 $\pm$ 0.18 s. In this time we were able to run 11$\pm$2 iterations of the MH algorithm.

The Metropolis Hastings algorithm is comparatively inefficient due to the matrix multiplication required to compute the forcing function $f$ from a given $q$ using the basis vectors (see equation \ref{eqn:reducedrank}). In the adjoint method this only needs to be calculated once, but is required at each step of the MH algorithm.

Whilst more sophisticated MCMC implementations will achieve faster convergence than random walk samplers, any MCMC algorithm will require 10,000s of  forward model evaluations to  approximate the same posterior the adjoint method can compute exactly\footnote{Here "exactly" refers to our ability to write down and compute the posterior for a given basis truncation of the GP, rather than approximating the posterior with MCMC.} in a fraction of the time.

\section{Discussion}
\label{sec:disc}
This work was motivated by the problem of inferring the distribution of spatially and temporally varying sources of pollution  across a city from noisy observations using a model of the pollution's atmospheric transport. Estimating the pollution concentration and its sources can help local authorities reduce the population's exposure, and to motivate policy interventions.
%Many related problems and various approaches exist. 
% As mentioned in the introduction, this challenge is known as the inverse problem: trying to find the sources from observations. Much of the earlier work in this field focused on inferring the location of a single pollution source \citep[see][for a review]{hutchinson2017review}. This was particularly relevant when determining the source of released radioisotopes \citep{pudykiewicz1998application} during nuclear tests or containment failures in reactors or storage facilities. Of more relevance to our work are approaches which seek to estimate the distribution of multiple pollution sources. Historically, a common approach was to divide the domain into source regions, run an atmospheric model of advection and diffusion and then optimise the values of the sources to find a MAP point estimate.
%The motivation for our paper is the problem of how to use noisy measurements of air pollution to infer the spatially and temporally varying source of the air pollution, using an advection-diffusion model for the spread of the pollution. 
Linear systems such as this, are typical of the type of challenge faced throughout the sciences and engineering, and still pose computational challenges that make principled statistical inference almost intractable, or if they are tractable, the computational cost (e.g. of using MCMC) is such that only a limited range of models and situations can be analysed.  %This paper demonstrate how the development of an adjoint model can lead to computational benefits in many situations. We have shown how conjugate Bayesian inference can be performed for an unknown GP forcing function at the cost of doing $n$ (the number of data points)  adjoint solves.
In this paper we developed an approach that results in conjugate Bayesian inference of an unknown GP forcing function, with a computational cost that scales  linearly with the number of observations, $n$.
As $n$ increases, the approach may eventually require more computation than competitor methods such as MCMC (which in theory has cost independent of $n$), but given that MCMC typically requires $10^5-10^6$ iterations even for low-dimensional problems\footnote{In the best case scenario of independent Gaussian posteriors, the number of iterations required for random walk Metropolis Hasting  scales as  $\mathcal{O}(M^{2})$ \citep{roberts1997} and $\mathcal{O}(M^\frac{5}{4})$ for Hamiltonian Monte Carlo,   \citep{Beskos2013} where $M$ is the dimension of the parameter, with each iteration requiring a solve of the forward problem. For more complex problems, scaling rates will be worse.},
there is still a large range of problems for which an adjoint may be beneficial. In our PDE example, the adjoint-aided approach required orders of  magnitude fewer PDE solves than MCMC. We should note that a key limitation of our approach is that it only applies to linear systems and cannot be used to determine forcing functions for non-linear systems such as the Navier-Stokes equations.

%The cost of our approach scales with the number of data points $n$. For Bayesian inference, MCMC algorithms typically use a minimum of $10^5$ iterations, even for low-dimensional problems. In cases where the dimension of the parameter is large, many more iterations are typically needed. Whilst it is the case that for big-$n$ problems, the use of an adjoint will lead to no benefit, the .......arghhhhh.

% MOVED - rewrite

% A second advantage of using adjoints is that the gradient  of a cost function depending on the solution $u$ can be computed with respect to parameters $p$, using the  {\it adjoint sensitivity} approach \citep[see, e.g.][]{Bradley2010,margossian2019,yashchuk2020}.
% For example, consider the unregularized sum of squares/log-likelihood $S(p,q) = (z-h(u))^\top (z-h(u))$ from Eq.~(\ref{eqn:sos}), where $u$ satisfies $\Lo_p u =f_q$. We've shown how we can estimate $\hat{q}(p)=\arg \max S(p,q)$ or find the posterior $\pi(q\mid z, p)$, but using the adjoint sensitivity approach  also gives us $\frac{\partial}{\partial p}S(p,q)$. This can be used either by a gradient-based optimizer to learn $\hat{p}=\arg \max S(p, \hat{q}(p))$, or in a variational inference scheme that requires gradients to estimate the variational parameters such as the variational autoencoder \citep{kingma2014,roeder2019}, or in a Hamiltonian Monte Carlo algorithm \citep{neal2011} to efficiently target the posterior (Eq.~\ref{eqn:Bayes0}). We do not explore these approaches in detail here.

% END_MOVED

We  only briefly touched upon the problem of estimating the  operator's non-linear parameters $p$ (e.g.,  advection and diffusion rates) and the GP hyperparameters.  {\it Adjoint sensitivity} methods \citep[see, e.g.][]{Bradley2010,margossian2019,yashchuk2020} can be used to estimate gradients of the log-likelihood with respect to the non-linear parameters in a numerically stable alternative to  automatic-differentiation frameworks (such as TensorFlow), which can be unstable when back propagating through long iterative loops. Access to gradient information allows inference of the additional parameters to be performed efficiently within the preferred statistical paradigm (e.g. maximum likelihood using Bayesian optimization - see the supplementary information; Hamiltonian Monte Carlo \citep{neal2011}; or a variational autoencoder \citep{kingma2014} framework etc.). %Note that adjoint systems are of the same type as the original system, so the knowledge and experience gained with the forward model will be of relevance to solving adjoint systems. 
%A cautionary note is that when the observation operator $\htild$ has small support, such as for pointwise evaluation (i.e., $h(u) = u(x,t)$),  care is needed when using numerical methods to solve the  adjoint systems, as adaptive step size methods can  miss small non-zero regions in $\htild$.

Finally, there are many ways in which this approach may be accelerated, for example, by the use of intelligent numerical solvers that reuse solution trajectories and  adaptive step size solvers, multi-fidelity methods that use varying grid sizes, and stochastic approaches which use only a subset of the data at each stage. The core computational tasks in our approach (solving the adjoint systems)  are embarrassingly parallelisable, enabling easy deployment on HPC facilities if required.
%We have found the main computational cost in our method, beyond the requirement to derive and implement an adjoint, is the computation of the dot product of the bases and the adjoints over the grid. In future work we will explore how this can be done far more quickly and with a smaller memory footprint.

% Although we can write the forward model in auto-diff framework (such as Tensorflow) in order to find  gradients with respect to these  parameters, the long iterative loops have been found to lead to numerical instabilities \citep{. 
% A potential benefit of using the adjoint approach is immediate access to these gradients....?? We can then optimise these using standard gradient descent (on the marginal log likelihood)??????

%Discuss using derivs to infer p, SGD, more efficient solvers (storing backwards runs upto points), 
%difficulty of adaptive step-sizes, multi-fidelity methods.

%\section*{Acknowledgements}

{\bf Software:} The algorithm has been implemented (for both the ODE and PDE problems) in a Python module available at https://github.com/SheffieldML/advectionGP. The repository also contains Jupyter notebooks that are used to produce the figures and tables in this paper.

\section*{Acknowledgements}

This work was directly funded by EPSRC projects EP/T00343X/1 and EP/T00343X/2. In addition, EB was supported by funding from Google.org and RW by EPSRC projects EP/P010741/1, EP/W000091/1, and EP/X012603/1.

\clearpage

% In the unusual situation where you want a paper to appear in the
% references without citing it in the main text, use \nocite

\bibliographystyle{icml2022}

\bibliography{Kampala_pollution}

%%%%%%%%%%%%%%%%%%%%%%%%%%%%%%%%%%%%%%%%%%%%%%%%%%%%%%%%%%%%
\clearpage
\appendix
\section*{Supplementary material for {\it Adjoint-aided inference of Gaussian process driven differential equations}}

\section*{Latent force models using Green's functions}

Existing approaches to latent force models rely upon the concept of a Green's function \citepsupp[e.g.][]{higdon2002space, boyle2004dependent, alvarez2009latent, alvarez2013latent, guarnizo2018latentrff}. Here, we briefly describe this approach and how it links to our adjoint-aided approach, and discuss the advantages and disadvantages of both methods. Consider the linear system
\begin{align}
\Lo u&= f \qquad \mbox{ for } x\in \Omega \label{eqn:System2}\\
u&=0 \qquad \mbox{ for }x\in \partial\Omega.\nonumber
\end{align}
Here, $\Lo$ is assumed to be a differential operator, and the solution $u$ is a function of $x$ with domain $\Omega$. The Green's function for this system, $G_y(x)$, satisfies 
\begin{align}
\La G_y(x)&= \delta_y(x) \qquad \mbox{ for } x\in \Omega \label{eqn:System3}\\
G_y(x)&=0 \qquad \qquad \mbox{for }x\in \partial\Omega\nonumber
\end{align}
where $\delta_y(x)=\delta(x-y)$ is the Dirac delta function, and $\La$ is the adjoint of $\Lo$. Once we have determined a Green's function, solution of the original problem (\ref{eqn:System2}) can be found by computing the convolution of $G$ with $f$:
% \begin{align*}
%     u(y) &= \langle \delta_y(\cdot), u(\cdot)\rangle = \langle \La G_y(\cdot), \; u(\cdot)\rangle \\
%     &= \langle G_y(\cdot), \; \Lo u(\cdot)\rangle = \langle G_y(\cdot), \; f(\cdot)\rangle\\
%     &= \int G_y(x) f(x) \rd x
% \end{align*}
\begin{align*}
    u(y) &= \langle \delta_y, u\rangle \qquad\qquad \mbox{ by definition of Dirac delta}\\
    &= \langle \La G_y, \; u\rangle \qquad \; \mbox{ by Eq. (\ref{eqn:System3})}\\
    &= \langle G_y, \; \Lo u\rangle \qquad \;\;\;\mbox{ by definition of the adjoint}\\
    &= \langle G_y, \; f\rangle \qquad\quad \;\;\mbox{ by Eq. (\ref{eqn:System2})}\\
    &= \int G_y(x) f(x) \rd x.
\end{align*}

The standard approach to latent force models then assumes $f$ is a Gaussian process, 
$f\sim GP(0, k),$
and uses the linearity of this expression and the closure of Gaussian processes under linear operations \citepsupp{rasmussen2003gaussian} to conclude that 
$u$ is also distributed as a Gaussian process,  
$$u \sim GP(0, k_u)$$ 
with covariance function 
\begin{equation}
k_u(y, y') = \int G_{y}(x) \int G_{y'}(x') k(x,x') \rd x' \rd x.\label{eqn:LFM}
\end{equation}
For some forms of the kernel $k$, e.g. the exponentiated quadratic kernel, it is possible to evaluate these integrals analytically when $G$ is known. Alternatively, we can resort to numerical integration to evaluate Eq.~(\ref{eqn:LFM}), for example, using random Fourier features \citepsupp{guarnizo2018latentrff}. Other works have represented $G$ using a polynomial series \citepsupp{GuarnizoLaguerre2018} or have put another GP prior over $G$ \citepsupp{tobar2015learning}.

When the Green's function is known for a given system, this approach can work efficiently and may perform as well or better than the adjoint-aided approach. See \citetsupp{Cole2000} for a comprehensive list of Green's functions. However, for many systems (particularly operators with spatially/temporally varying coefficients) the Green's functions are not analytically computable. For diagonalizable operators, we can try to find  the eigenfunctions of $\Lo$, i.e.,  $\lambda_i$, and $\phi_i(x)$ such that  $\Lo \phi_i = \lambda_i \phi_i$,  then we can write
$$G_y(x) = \sum_{i=1}^\infty \frac{1}{\lambda_i}\phi_i(x) \phi_i(x').$$
If we can estimate $\lambda_i$ and $\phi_i(x)$ numerically (i.e., by numerically solving the differential equation  $\Lo \phi_i = \lambda_i \phi_i$ on some computational mesh) we can then truncate this sum and form a numerical approximation of $G_y(x)$. But in this case, we would then need to use our numerical approximation of $G$ in a further numerical approximation of the integral in Eq.\~(\ref{eqn:LFM}) which can easily lead to numerical instabilities and low accuracy. In addition, not all differential operators  are diagonalizable (i.e., admit a basis of eigenfunctions), for example,  operators which are not self-adjoint.

In contrast, our adjoint-aided approach relies solely on the existence of the adjoint operator $\La$ and our ability to solve adjoint systems numerically. To do this, we can deploy modern finite element solvers that are efficient, stable, and offer good error-control. A full numerical analysis of the respective errors of the two approaches is beyond the scope of this paper, and  would necessarily be  specialized to the implementation details of all the particular numerical algorithms used. 

In summary, we would recommend that in the special case where $G$ is known and Eq.~(\ref{eqn:LFM}) is tractable, that a Green's function approach be used. In other situations, the ease of the adjoint approach introduced here is likely to be an attractive alternative both in terms of accuracy, numerical stability, and ease of implementation.

\section*{Comparison to competing methods}

We conducted a comparison between the adjoint method, the Green's function method and a classical Gaussian process on the ordinary differential equation model presented in section 4.1. Observations were taken at 20 time points over $t\in[0,10]$ with a grid resolution of 200.
    \begin{itemize}
        \item The Gaussian process had a mean squared error of 0.0055 between the true output and the inferred output.
        \item The Green's function method (as in \citetsupp{alvarez2009latent}) had an MSE of 0.0051 for the output error and 0.0860 for the source error.
        \item The Green's function method with random Fourier features (as in \citetsupp{guarnizo2018latentrff}) achieved the following MSEs:
        \begin{itemize}
            \item 20 features: Source MSE of 0.099 and output MSE of 0.0058
            \item 200 features: Source MSE of 0.0927 and output MSE of 0.0055.
            \item 500 features: Source MSE of 0.0856 and output MSE 0f 0.0052.
            \item 2000 features: Source MSE of 0.0861 and outpute MSE of 0.0051.
        \end{itemize}
        \item The adjoint method with M=2000 random Fourier features had an MSE of 0.0056 between the ground truth concentration and the inferred concentration and an MSE of 0.079 between the ground truth and inferred sources.
    \end{itemize}
   All three methods achieve a similar quality of inference over the system output. This is to be expected as all three methods utilise a similar statistical model. For larger numbers of features ($M\sim200$) the adjoint method and the GP predicted the system response with similar accuracy. It should be noted that by using a classical GP approach it is not possible to infer the unknown forcing function, $f$, which is one of the key advantages of the adjoint method. The Green's function method also performs to a similar level of accuracy as the adjoint method, though the Green's function method with Fourier features appears to perform better at low numbers of features for this particular test case.

\section*{Derivation of the Advection Diffusion Adjoint Equation}

Consider the advection diffusion operator discussed in Section 4.2:
\begin{equation}
\Lo u=\frac{\partial u}{\partial t} + \bp_1\cdot\nabla u -\nabla \cdot (p_2 \nabla u) \mbox{ in } \mathcal{X}\times [0,T]
\label{eqn:advec2}
\end{equation}
with initial  condition
\begin{equation}
u(x,0)=0 \ \mbox{ for all }\ x  \in \mathcal{X} \label{eqn:bcs4}
\end{equation}
and Neumann boundary condition
\begin{equation}
\nabla_{n}u=0 \ \mathrm{for \ x}  \in\partial \mathcal{X},  \label{eqn:BC2}
\end{equation} 
where $\partial \mathcal{X}$ is the boundary of $\mathcal{X}$, $\nabla_n u =\nabla u \cdot \mathbf{\hat{n}}$ denotes the normal derivative of $u$, with $\mathbf{\hat{n}}(x)$ the outward facing normal of $\partial \mathcal{X}$ at $x$.
Let  $\Omega=\mathcal{X}\times [0,T]$ denote the spatial temporal domain of $u$.

The adjoint of the system defined by Eqs~(\ref{eqn:advec2}--\ref{eqn:BC2}) will depend on both the differential operator, and the specific initial and boundary conditions imposed. To derive this, we need to find a linear operator $\La$ and a set of boundary conditions so that the bilinear identity
$$\langle \Lo u, v\rangle = \langle u, \La v\rangle$$ 
is satisfied for all sufficiently smooth functions $u$ and $v$ with compact support in $\Omega$.
Let $v$ be such a function, and consider 
\begin{equation}
    \langle  \Lo u, v\rangle = \int_\Omega \left(\frac{\partial u}{\partial t} + \bp_1\cdot\nabla u - \nabla \cdot (p_2 \nabla u)\right) v\,\rd\Omega.
    \label{eqn:Inneruv}
\end{equation}
In the derivation below, we'll assume $\bp_1$ and $p_2$ are constant, and follow the general steps outlined in \citetsupp{estep2004short}. As in the ODE example, the derivation essentially relies upon repeated application of integration by parts. 
For the first term in Eq.~(\ref{eqn:Inneruv}):
\begin{align*}
  \int_\Omega \frac{\partial u}{\partial t}v\,\rd\Omega &= \int_\Omega 
  \frac{\partial }{\partial t}(uv)- u\frac{\partial v}{\partial t}\,\rd\Omega\\
  &=\int_\mathcal{X}\int_0^T \frac{\partial }{\partial t}(uv)\,\rd t\,\rd x  -
  \int_\Omega u\frac{\partial v}{\partial t}\,\rd\Omega\\
  &= \int_\mathcal{X}u(x,T)v(x,T)-u(x,0)v(x,0)\,\rd x - \int_\Omega u\frac{\partial v}{\partial t}\,\rd\Omega.
\end{align*}
For the second term in Eq.~(\ref{eqn:Inneruv}):
\begin{align*}
    \bp_1\cdot\int_\Omega v\nabla u &=    \bp_1\cdot \left(\int_\Omega \nabla (uv) \,\rd \Omega- \int_\Omega u\nabla v \,\rd \Omega\right)\\
    &= \bp_1\cdot \left(\int_0^T \int_\mathcal{X} \nabla(uv) \,\rd x \,\rd t-\int_\Omega u\nabla v\,\rd \Omega
    \right)\\
&=   \bp_1\cdot\left(\int_0^T\oint_{\partial \mathcal{X}} uv \mathbf{\hat{n}} \,\rd x\; \,\rd t - \int_\Omega u\nabla v\,\rd\Omega\right)
\end{align*}
where the first equality uses the vector product rule, and the third the divergence theorem. For the third term in Eq.~(\ref{eqn:Inneruv}) we have
\begin{equation*}
p_2  \int_\Omega v\nabla \cdot \nabla u\,\rd\Omega = p_2\left(\int_0^T\oint_{\partial \mathcal{X}} v\nabla u \cdot \mathbf{\hat{n}} \,\rd x\,\rd t - \int_\Omega  \nabla v \cdot \nabla u \,\rd\Omega\right).
\end{equation*}
We can then repeat this process on the final term above 
$$\int_\Omega  \nabla v \cdot \nabla u \,\rd\Omega = \int_0^T\oint_{\partial \mathcal{X}} u \nabla v\cdot \bn \,\rd x \,\rd t - \int_\Omega u\nabla\cdot\nabla v \,\rd \Omega.$$

Combining all of these terms together gives
\begin{align*}
\langle  \Lo u, v\rangle&= \int_\Omega \left(- \frac{\partial v}{\partial t} - \bp_1\cdot\nabla v - \nabla\cdot (p_2\nabla v)\right) u \,\rd\Omega  \\ 
&\qquad +\int_\mathcal{X}u(x,T)v(x,T)-u(x,0)v(x,0)\,\rd x\\
 &\qquad + \int_0^T\oint_{\partial \mathcal{X}} uv\bp_1.\mathbf{\hat{n}} - p_2v \nabla u \cdot \mathbf{\hat{n}} + p_2u \nabla v\cdot \mathbf{\hat{n}} \,\rd x\,\rd t \\ 
&= \langle   u, \La v\rangle + \mathrm{ boundary \ terms.}
\end{align*}
As in the ODE case, we then choose the boundary and initial conditions on $v$ to make the boundary terms above vanish. 
Firstly, as $u(x,0)=0$ for all $x$,  setting the final condition $v(x,T)=0$ for all $x$  eliminates the first boundary term. Secondly, as $\nabla u \cdot \mathbf{\hat{n}}$ is  0 on the boundary (from the boundary conditions on $u$, Eq.~\ref{eqn:bcs4}), the third term also vanishes. Finally, to set the remainder of the boundary integral to 0 we assume  $\bp_1 v+p_2\nabla v=0$ for all $x\in \partial \mathcal{X}$. 

Thus our adjoint operator is  
\begin{equation}
    \La v = -\frac{\partial v}{\partial t} - p_1\nabla\cdot v - \nabla\cdot (p_2\nabla v)
\end{equation}
with final condition 
$$
v(x,T)=0 \mbox{ for all } x \in \mathcal{X}
$$
and mixed condition 
$$
\bp_1v+p_2\nabla v=0 \mbox{ for } x \in \partial \mathcal{X}.
$$

Note that when solving the original and adjoint systems numerically, checking that the bilinear identity does indeed hold is a useful validation of the derivation and PDE solvers.

\section*{PDE Inference Examples}

\begin{figure}[ht]
\begin{center}
\includegraphics[height=3.5cm]{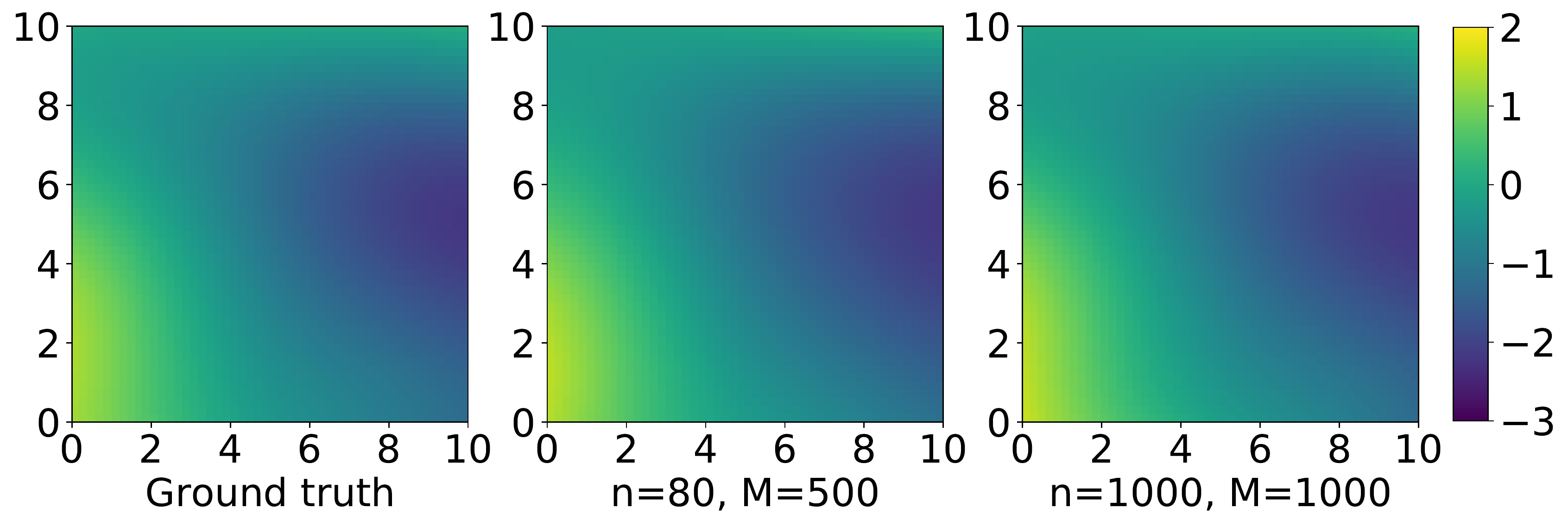}
\end{center}
\caption{Ground truth and posterior mean of the source when using kernel length-scale, $l=5$ (time-slice at $t=5$). The left image shows the ground truth source, the middle image shows the posterior mean inferred using 80 observations and 500 features (MSE=$0.008$), the right shows the posterior mean inferred using 1000 observations and 1000 features (MSE$=0.004$).}
\label{fig:compL5}
\end{figure}

Various factors effect the quality of source inference when using the adjoint method. These include the number of random Fourier features (RFFs), $M$, the number of observations, $n$, the locations of the sensors, and the ratio of the ground truth source lengthscale, $\lambda$, to the size of the domain. In Section 4.2 we investigated the effect of changing the values of $n$ and $M$ (see Table 2) for a system with a fixed lengthscale. Here we briefly illustrate the effect of changing $\lambda$. We consider a $10 \times 10 \times 10$ grid in space and time and two scenarios:
\begin{enumerate}
\item  100 sensors arranged in a grid, with readings at 10 points in time, using 1000 RFFs to infer the source ($n=1000$, $M=1000$); 
\item  16 sensors arranged in a grid, with readings at 5 time points, using 500 RFFs ($n=80$, $M=500$). 
\end{enumerate}
 We generated three ground truth sources using length-scales $l=5$, 2 and 1. In each case, we used the adjoint method in the scenarios described to infer the posterior distribution of the source. Fig.~\ref{fig:compL5} shows the ground truth generated with $l=5$ and the inferred source in each scenario at a single time-slice. In this case both models perform similarly on visual inspection. The MSE between the inferred source and the ground truth is $0.004$ for scenario 1 ($n=1000$, $M=1000$), and 0.008 in scenario 2 ($n=80, M=500$). 
 
\begin{figure}[ht]
\begin{center}
\includegraphics[height=3.5cm]{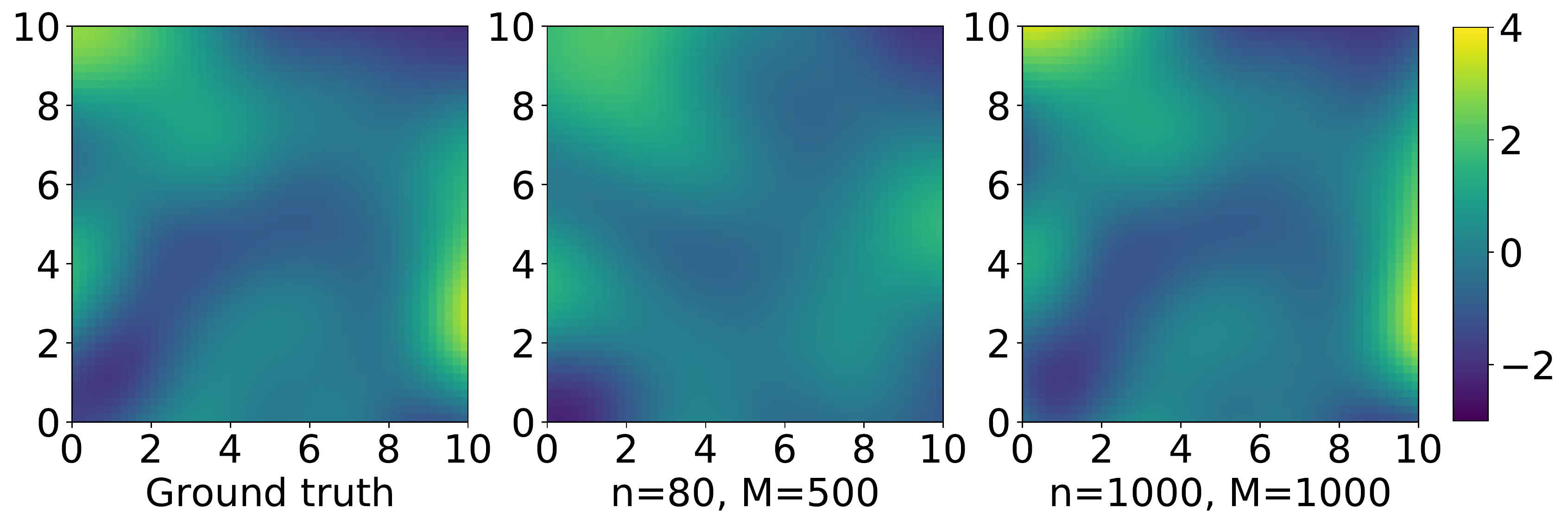}
\end{center}
\caption{Ground truth and posterior mean of the source when using  kernel length-scale, $l=2$ (time-slice at $t=5$). The left image shows the ground truth source, the middle image shows the posterior mean inferred using 80 observations and 500 features (MSE=$0.68$), the right shows the posterior mean inferred using 1000 observations and 1000 features (MSE=$0.07$).}
\label{fig:compL2}
\end{figure}

\begin{figure}[h!]
\begin{center}
\includegraphics[height=3.5cm]{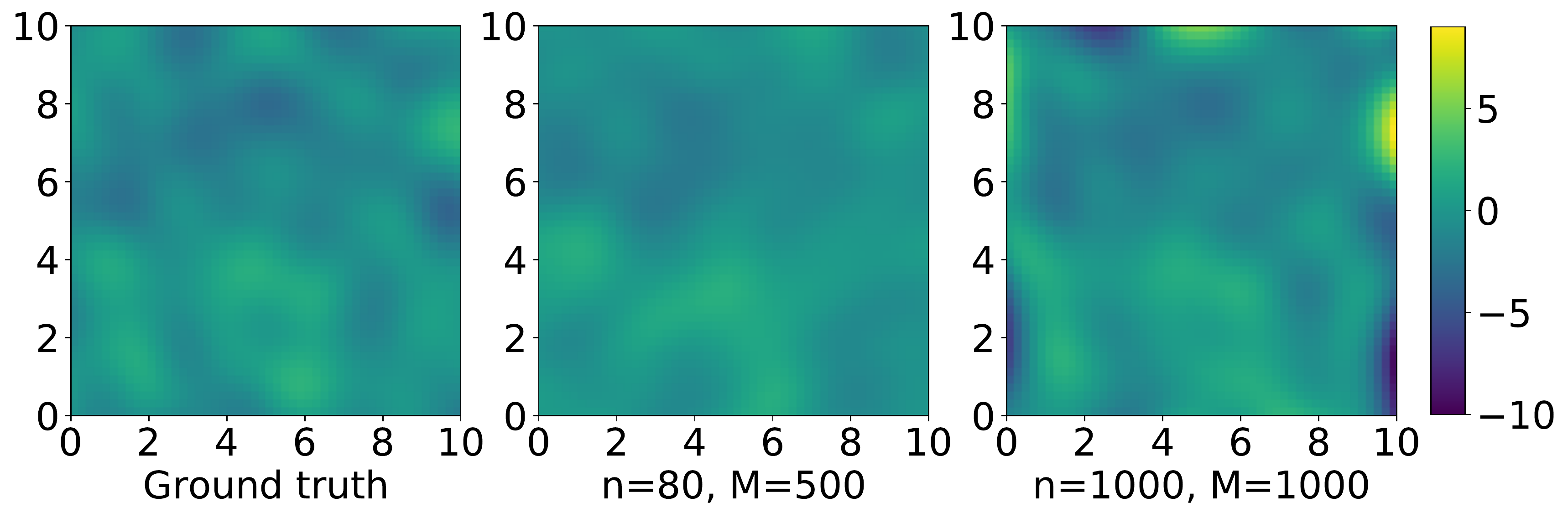}
\end{center}
\caption{Ground truth and posterior mean of the source from with kernel length-scale, $l=1$ (time-slice at $t=5$). The left image shows the ground truth source, the middle image shows the posterior mean inferred using 80 observations and 500 features (MSE=$1.85$), the right shows the posterior mean inferred using 1000 observations and 1000 features (MSE=$2.55$).}
\label{fig:compL1}
\end{figure}

Fig.~\ref{fig:compL2} shows the same information for the source generated with length-scale $l=2$. In this case, we can visually see that the posterior inference is much more accurate in scenario 1. The MSEs are 0.07 for scenario 1 and 0.68 for scenario 2. 
Finally, in the case where $l=1$ (see Fig.~\ref{fig:compL1}), visual inspection reveals that in scenario 2, the posterior mean bears little resemblance to  the ground truth,  whereas key features of the ground truth are visible in the posterior mean for scenario 1. This is reflected in the MSEs, which are 1.85 for scenario 1 ($n=1000$) and 2.55 for scenario 2 ($n=80$).

These results demonstrate  the expected phenomena, namely that  as the ratio of the length-scale to the grid size decreases, more features and observations are required to accurately infer the ground truth. Furthermore, in the short length-scale case the accuracy of inference is generally lower,  as the source varies more between sensor locations than in the longer length-scale case. Additionally, in longer length-scale cases, fewer features and observations are required for high quality inference, thus enabling inference with less computational resource.

\begin{figure}[h!]
\begin{center}
\includegraphics[width=6cm,trim=0 1cm 0 0]{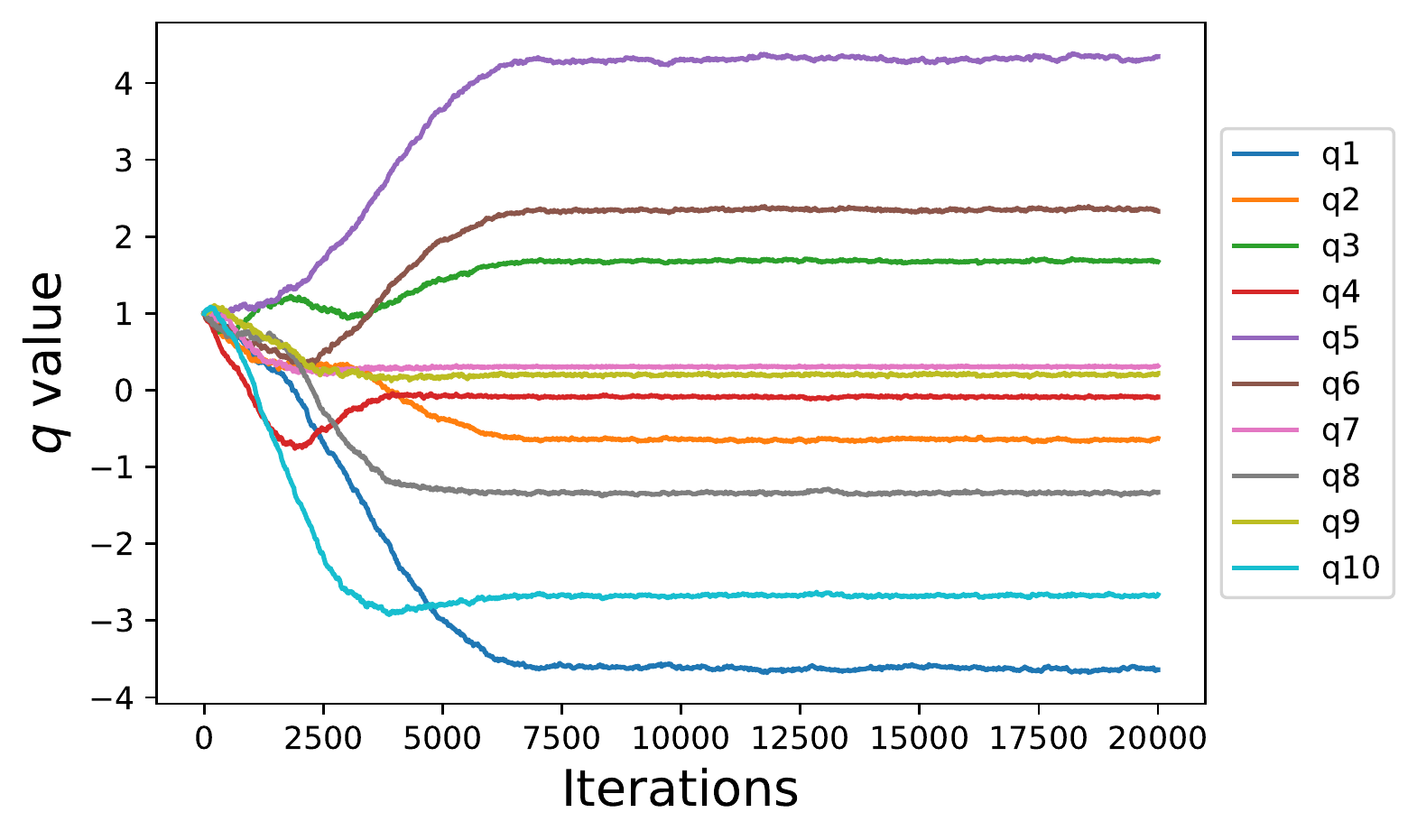}
\end{center}
\caption{Trace plots for MCMC corresponding to Table \ref{tab:MHvAdj}}
\label{fig:MCMC}
\end{figure}

Finally, Fig.~\ref{fig:MCMC} shows the trace plot for the $q$ parameters for an implementation of the Metropolis Hastings algorithm in the case where $M=10$ features  are used. See the main text (Sect. 4.2) for details.

\section*{Bayesian Optimisation Output}

Although not our focus here, we note that we can infer the remaining parameters $\bp_1$, $p_2$ (PDE) and $\tau^2, \lambda$ (GP hyperparameters) in a variety of ways. By way of illustration, here we use %we embedded our method into a Bayesian optimisation framework using the GPyOpt package. 
the GPyOpt \citepsupp{gpyopt2016} package to maximise  the negative log-likelihood using the expected improvement acquisition function in a 
Bayesian optimization approach \citepsupp{shahriari2015taking}. 

To perform inference for $\bp_1, p_2, \tau^2, \lambda$, we wrote a function which, given a parameter array $\theta$, estimates the posterior predictive accuracy of our source posterior when  using $\theta$ (here $\theta$ may contain a subset of the parameters). We do this by first simulating 100 realizations from the posterior mean of the source for a fixed parameter, i.e., $f_1, \ldots, f_{100}\sim p(f | z, \theta)$, and we push these through the PDE to get a posterior predictive sample of concentration fields $u_1, \ldots, u_{100} \sim p(u |z, \theta)$. The  negative log-likelihood is then calculated between these and the training observations giving us a way to score parameter $\theta$. The negative log-likelihood function was used as the objective function in a Bayesian Optimisation routine \citetsupp{gpyopt2016}. To test this approach we generated various ground truth source and solution fields using fixed values of $\theta$.

\begin{figure}[ht]
\includegraphics[width=\columnwidth]{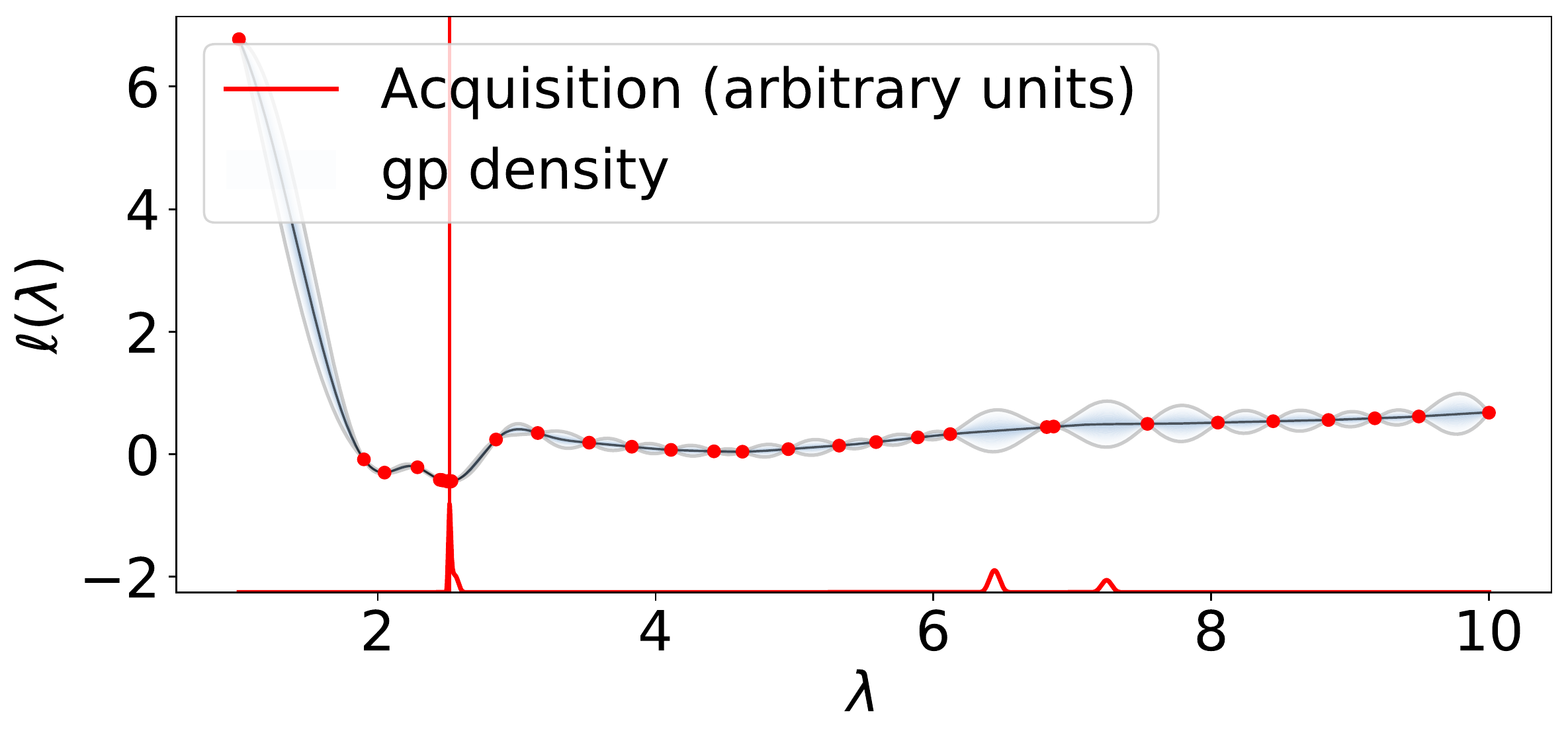}
\caption{The output of the Bayesian Optimisation algorithm used to infer the value of the GP kernel length-scale, $\lambda$. In this case the true value of $\lambda$ is 2 and the algorithm found the minimum of the negative log-likelihood at $\lambda = 2.52$. This plot shows where the objective function was evaluated, and the posterior mean and variance at each point.}
\label{fig:FindLength}
\end{figure}

Fig.~\ref{fig:FindLength} shows the exploration and eventual convergence in a particular case where we used  $\lambda=2$ to generate a ground truth source. In this case, the maximum likelihood estimate of $\lambda$ was found to be $\hat{\lambda}=2.52$ which the optimization found after 29 iterations.  In a case where the true kernel length-scale and variance were both 2, and the wind-speed was 0.04, Bayesian optimisation found the maximum likelihood values of 1.24, 3.20 and 0.031 respectively after 20 iterations. 
Further work is needed to fully explore how to embed this approach into parameter estimation schemes, but we hope it gives some insight into how parameter estimation could be performed. Finally, note that the adjoint approach gives  the possibility of estimating the gradient of the loss function. This would enable the adjoint approach to be embedded into  gradient based inference algorithms such as the VAE and Hamiltonian Monte Carlo, hopefully accelerating inference.

%SAY we could just as easily do VI, MCNC etc.

% MCMC for ODE?

\section*{PDE model applied to the Round Hill II dataset}

\begin{figure}[h]
\includegraphics[width=\columnwidth]{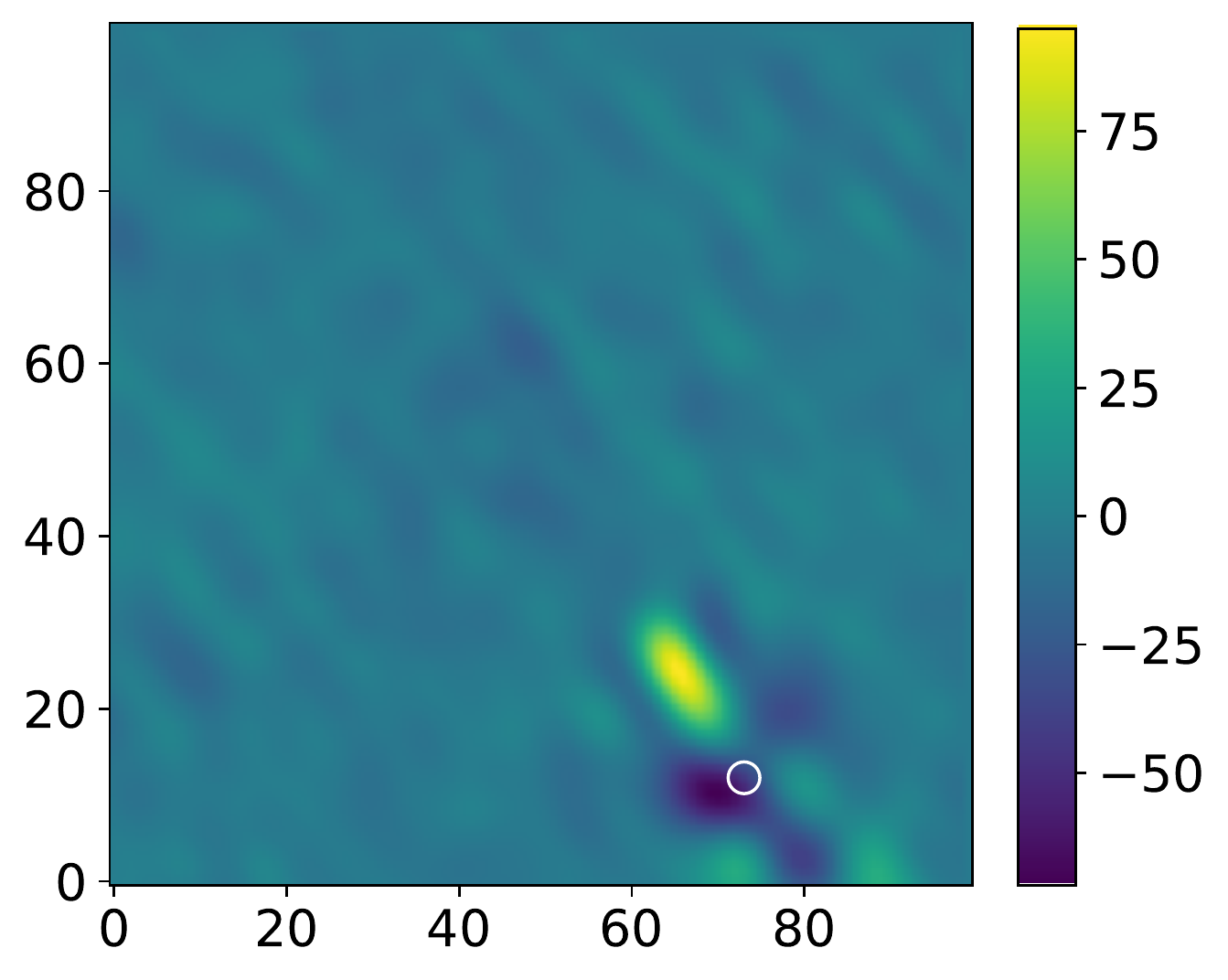}
\caption{The mean inferred source in the Roundhill experiment at $t=0$. The white circle indicates the true source location.}
\label{fig:RoundhillSourceInfer}
\end{figure}

\begin{figure}[h]
     \centering
     \begin{subfigure}{0.45\columnwidth}
         \centering
         \includegraphics[width=\columnwidth]{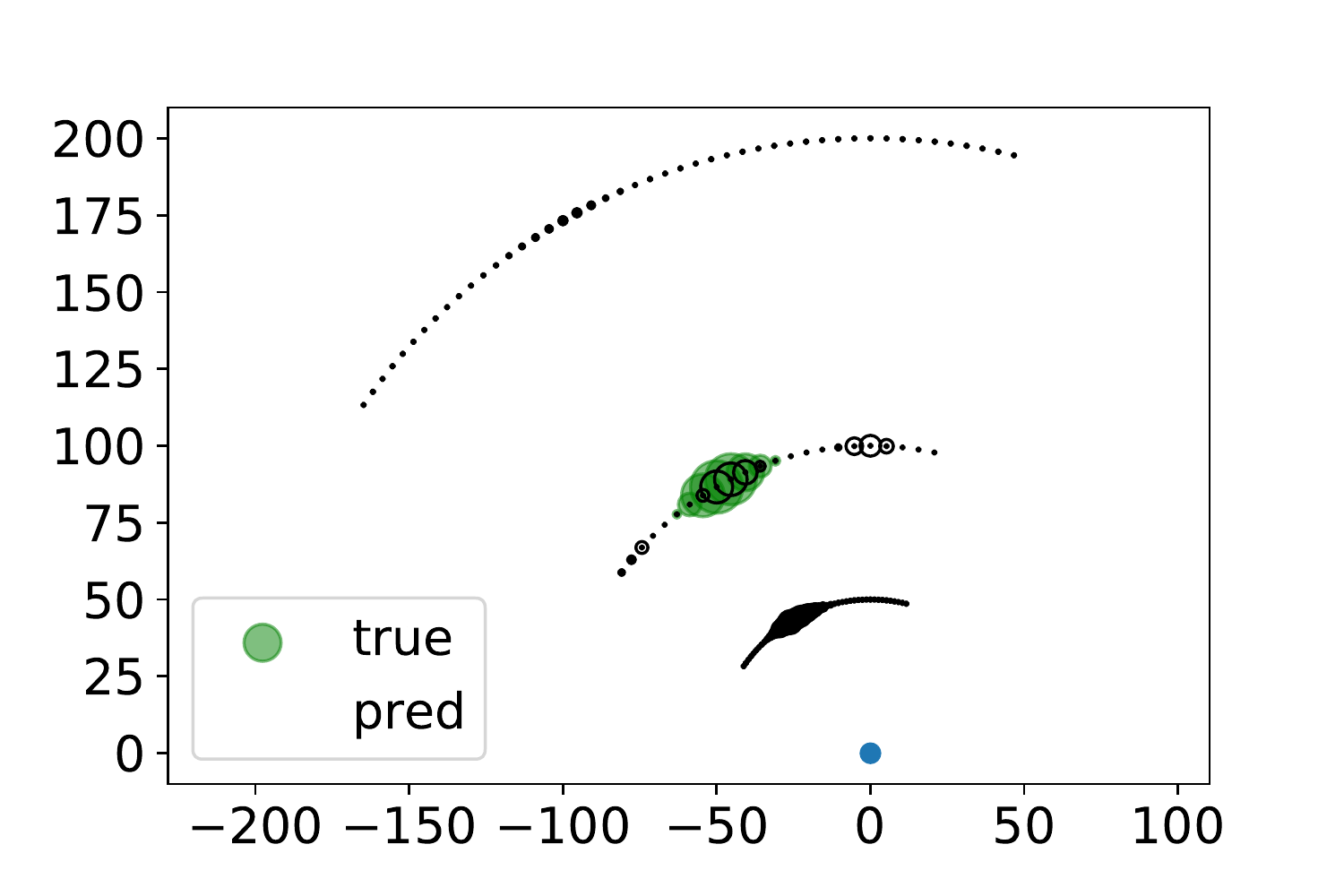}
         \caption{The relative pollutant concentrations in the Roundhill experiment between the predicted concentration inferred using the adjoint method and the true values.}
         \label{fig:RoundhillAdjointObservation}
     \end{subfigure}
     \hfill
     \begin{subfigure}{0.45\columnwidth}
         \centering
         \includegraphics[width=\columnwidth]{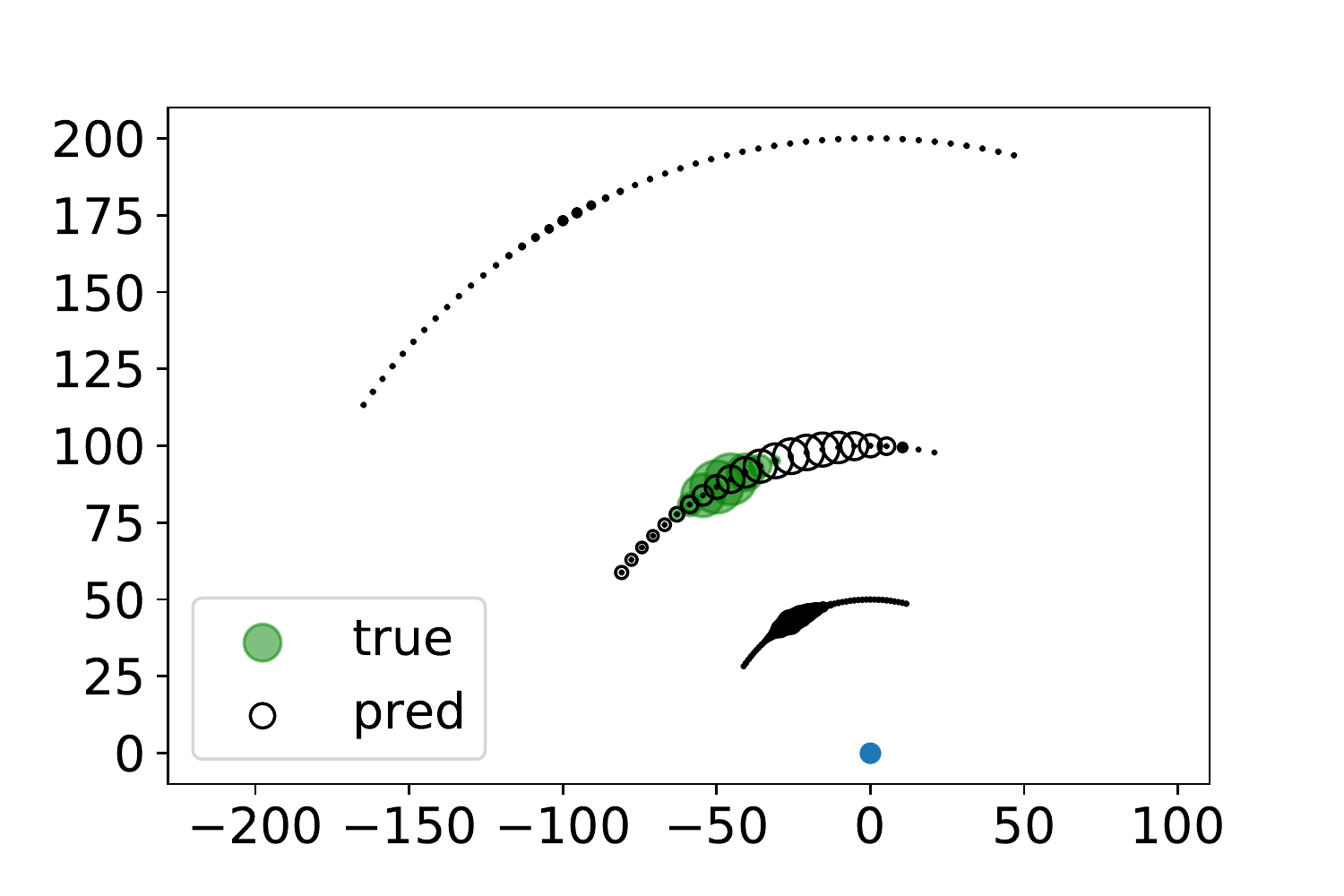}
         \caption{The relative pollutant concentrations in the Roundhill experiment between the predicted concentration inferred using a Gaussian process and the true values.}
         \label{fig:RoundhillGPObservation}
     \end{subfigure}
     \caption{}
     \label{fig:RoundhillObservations}
\end{figure}

To test the approach using data from a physical experiment, we used the Round Hill II advection-diffusion experiment \citepsupp{cramer1957field}. In this study, researchers deployed 183 midget impingers for measuring sulphur dioxide in three partial concentric rings, 50m, 100m and 200m downwind from the release site, spanning $69^\circ$. A constant source of sulphur dioxide (releasing approximately 5-10 $g s^{-1}$) was used over a ten minute period, during which the impingers took measurements of average concentration over the first 30 seconds, 3 minutes and 10 minutes. The average wind speed and direction was recorded (2.14 $m s^{-1}$). We modelled this with our adjoint approach, as in section 4.2, over a $250m \times 250m$ domain spanning 13 minutes, using 10,000 random Fourier bases to approximate the Gaussian process forcing term. We tested two aspects of our model's capabilities. First: Source attribution. The model's mean source prediction was roughly flat except for a peak approximately 45m downwind of the true release site, see Figure \ref{fig:RoundhillSourceInfer}. This discrepancy is expected as the true dataset contained a point source while our model had a GP prior (with EQ kernel and lengthscale of 10m) over the source. This leads to an inferred broader source, slightly closer to the ring of sensors. The second test was predicting the SO${}_2$ concentration: We removed the middle (100m) ring of sensors from the training data, then tried to predict their measurements. For comparison we used a Gaussian process with a length-scale of 30m (and 30s) to predict the concentration. We found it useful to threshold the concentrations to be non-negative for both methods. Our model performed considerably better than the GP model. For the three measurement periods the results were:
    \begin{verbatim}
         Our Model    GP Model
    30s   14444        21385
    180s   6628        12968
    600s   4503         8490
    \end{verbatim}
    [measurement units were $mg/m^3$, so these MSEs are in $(mg/m^3)^2$ ]
See figure \ref{fig:RoundhillObservations} for a comparison of the inferred concentrations between the two models. Figure \ref{fig:RoundhillGPObservation} indicates that the Gaussian process generally overestimated the right hand side of the left out sensor array, whereas it can be seen from \ref{fig:RoundhillAdjointObservation} that the adjoint method predicted the true high concentration area fairly well, with some smaller overprediction at the left and right hand sides of the array.
    
\section*{Shift operators}

Here we provide an example of the adjoint method applied to a non-differential operator: the shift operator. Consider the operator $\Lo_a: \mathbb{R} \rightarrow \mathbb{R}$ such that
\begin{equation}
    \Lo_a(u(t))=u(t+a).
\end{equation}
This is the right shift operator. We can derive the adjoint of the right shift operator by taking the following inner product and using a change of variable (x=t+a).
\begin{equation}
    \langle \Lo u, v\rangle = \int_{-\infty}^{\infty} u(t+a)v(t)dt = \int_{-\infty}^{\infty} u(x)v(x-a)dx
\end{equation}
and so the adjoint of the right shift operator is the left shift operator, $\Lo^*_a: \mathbb{R} \rightarrow \mathbb{R}$, where $\Lo^*_av(t)=v(t-a)$. Having derived the adjoint of the right shift operator, it is possible to apply the adjoint method to an example system
\begin{equation}
    \Lo_a u(t) = u(t+a) = f(t) \label{eqn:shift}
\end{equation} where $f(t)$ is an unknown forcing function and observations of $u$ are obtained as noisy averages over short time windows (see equation \ref{eqn:odenoise}). 

\begin{figure}[ht]
     \centering
     \begin{subfigure}{0.45\columnwidth}
         \centering
         \includegraphics[width=\columnwidth]{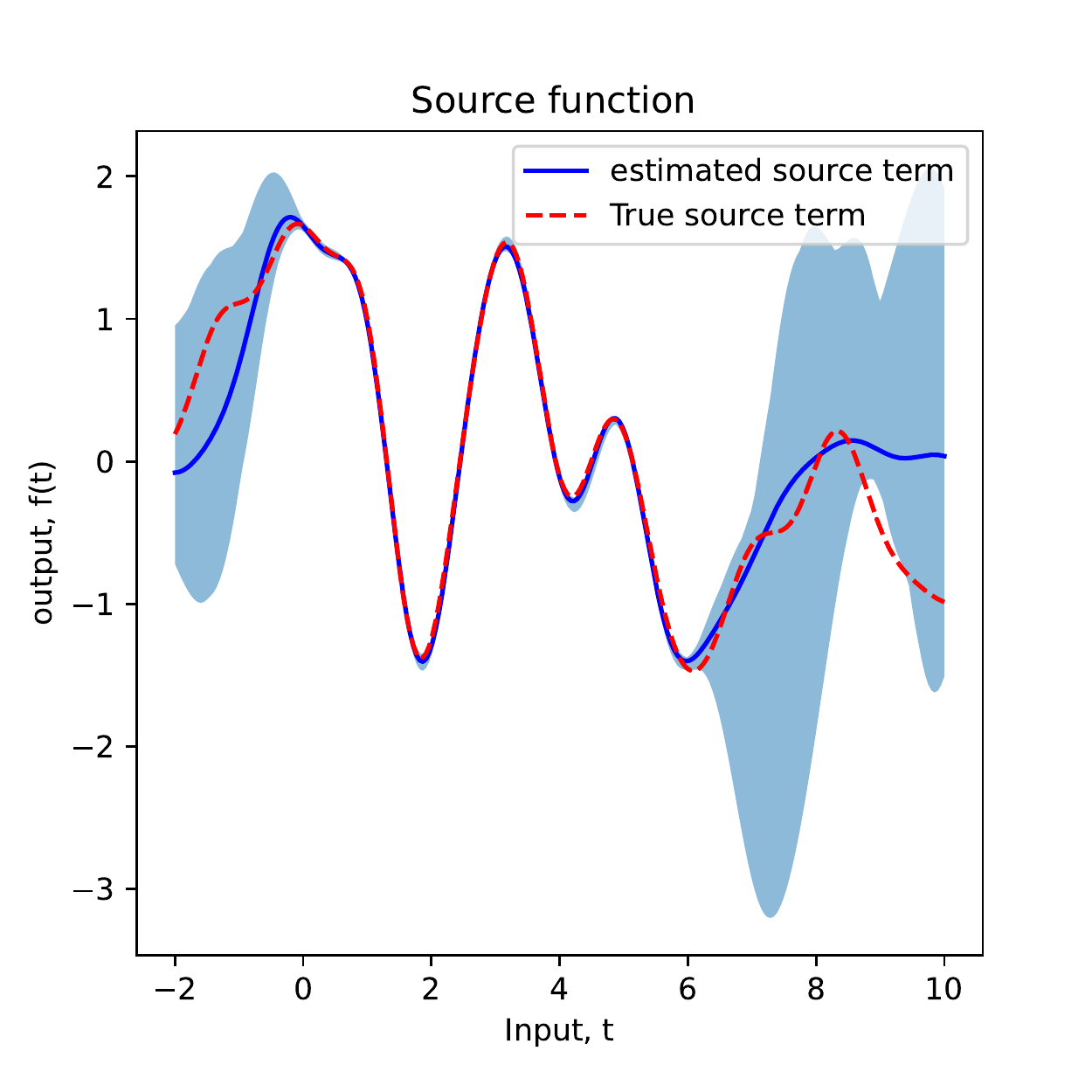}
         \caption{Inferred and ground truth source in the shift operator system with $a=2$}
         \label{fig:shiftSourcea2}
     \end{subfigure}
     \hfill
     \begin{subfigure}{0.45\columnwidth}
         \centering
         \includegraphics[width=\columnwidth]{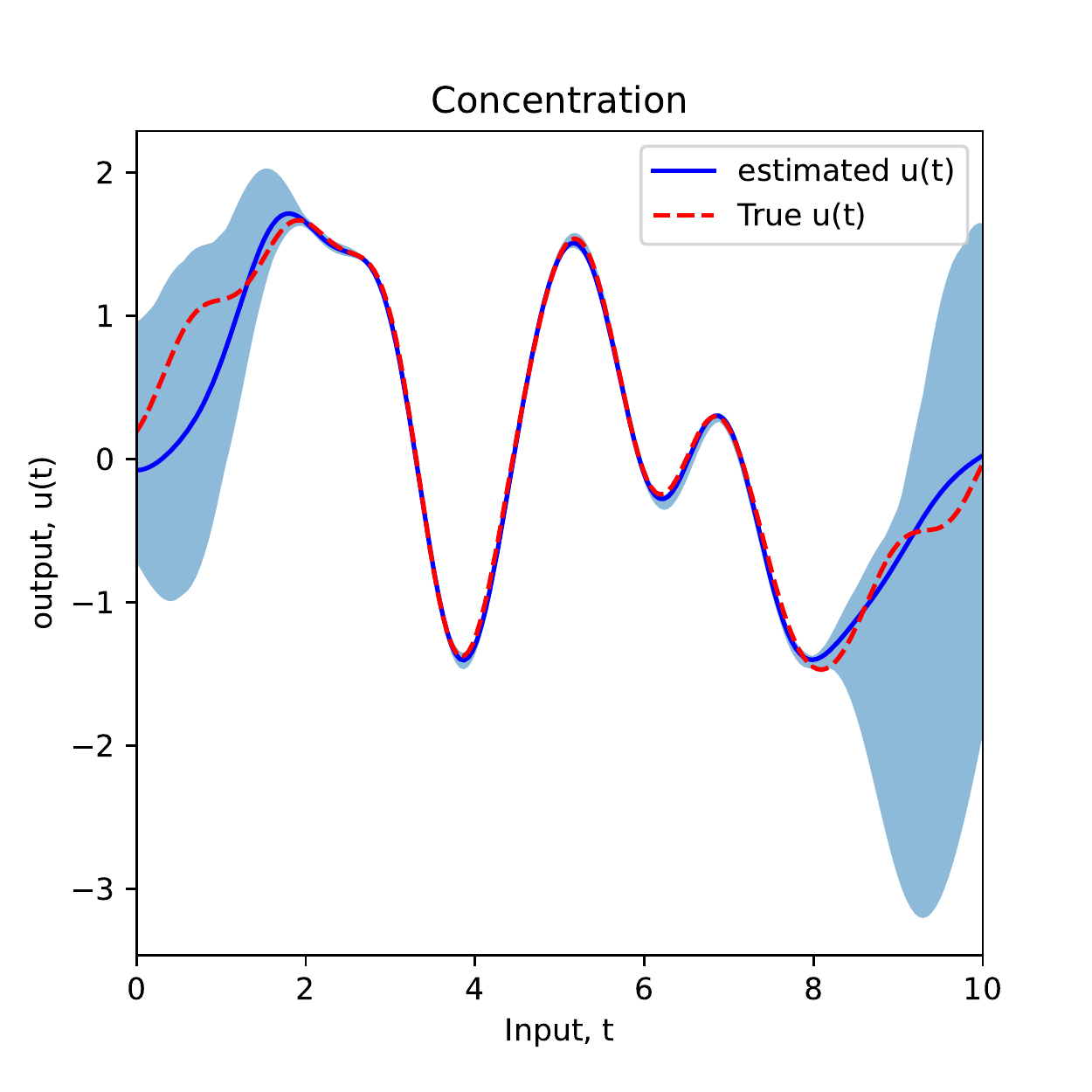}
         \caption{Inferred and ground truth output, $u(t)$ in the shift operator system with $a=2$}
         \label{fig:shiftOuta2}
     \end{subfigure}
     \caption{}
     \label{fig:shiftExample}
\end{figure}

Figure \ref{fig:shiftExample} shows the inferred and true $f(t)$ and $u(t)$ of the system given in equation \ref{eqn:shift} with $a=2$ and 20 observations evenly spaced between $t=2$ and $t=8$. Observations were generated with Gaussian noise, $\epsilon \sim N(0,0.05)$. The output $u(t)$ is well predicted within the observation range with relatively high certainty, the prediction is uncertain outside of the observation range. The forcing function, $f$ is well predicted between $t=0$ and $t=6$, i.e. the observation range shifted left by $a=2$. The MSE between the observations and the mean value of $u(t)$ inferred at the observations points is 0.003. For reference, the standard deviation of the observations is 1.001.  

It seems evident that to predict the source and output more confidently over the entire real line would require observations over the entire real line. Furthermore, it does not seem possible to infer the shift parameter, $a$. For example, if the true shift parameter is $a^*$ and the the model used to infer the source assumes shift parameter $a$, the inferred source will simply be shifted left by $a$ and the quality of prediction of observation points would be indistinguishable (given a fixed basis for expansion of $f$).    

\bibliographystylesupp{icml2022}
\bibliographysupp{Kampala_pollution}

\end{document}